\documentclass[sigconf]{aamas} 
\usepackage{balance} % for balancing columns on the final page
\setcopyright{ifaamas}
\acmConference[AAMAS '23]{Proc.\@ of the 22nd International Conference
on Autonomous Agents and Multiagent Systems (AAMAS 2023)}{May 29 -- June 2, 2023}
{London, United Kingdom}{A.~Ricci, W.~Yeoh, N.~Agmon, B.~An (eds.)}
\copyrightyear{2023}
\acmYear{2023}
\acmDOI{}
\acmPrice{}
\acmISBN{}

\acmSubmissionID{561}

\title[GANterfactual-RL]{GANterfactual-RL: Understanding Reinforcement Learning Agents' Strategies through Visual Counterfactual Explanations
}

\author{Tobias Huber}
\affiliation{
  \institution{University of Augsburg}
  \city{Augsburg}
  \country{Germany}
  }
\email{tobias.huber@uni-a.de}

\author{Maximilian Demmler}
\affiliation{
  \institution{University of Augsburg}
  \city{Augsburg}
  \country{Germany}
  }
\email{maxdemmler@googlemail.com}

\author{Silvan Mertes}
\affiliation{
  \institution{University of Augsburg}
  \city{Augsburg}
  \country{Germany}
  }
\email{silvan.mertes@uni-a.de}

\author{Matthew L. Olson}
\affiliation{
  \institution{Oregon State University}
  \city{Corvallis, OR}
  \country{United States}
  }
\email{olsomatt@oregonstate.edu}

\author{Elisabeth André}
\affiliation{
  \institution{University of Augsburg}
  \city{Augsburg}
  \country{Germany}
  }
\email{andre@informatik.uni-augsburg.de}

\begin{abstract}
Counterfactual explanations are a common tool to explain artificial intelligence models. For Reinforcement Learning (RL) agents, they answer "Why not?" or "What if?" questions by illustrating what minimal change to a state is needed such that an agent chooses a different action. Generating counterfactual explanations for RL agents with visual input is especially challenging because of their large state spaces and because their decisions are part of an overarching policy, which includes long-term decision-making. However, research focusing on counterfactual explanations, specifically for RL agents with visual input, is scarce and does not go beyond identifying defective agents. It is unclear whether counterfactual explanations are still helpful for more complex tasks like analyzing the learned strategies of different agents or choosing a fitting agent for a specific task. We propose a novel but simple method to generate counterfactual explanations for RL agents by formulating the problem as a domain transfer problem which allows the use of adversarial learning techniques like StarGAN. Our method is fully model-agnostic and we demonstrate that it outperforms the only previous method in several computational metrics. Furthermore, we show in a user study that our method performs best when analyzing which strategies different agents pursue.
\end{abstract}


\newcommand{\BibTeX}{\rm B\kern-.05em{\sc i\kern-.025em b}\kern-.08em\TeX}

\newcommand{\control}{Control}
\newcommand{\olson}{CSE} % Wasserstein-CSE, W-CSE
\newcommand{\starGAN}{GANterfactual-RL}
\newcommand{\blueGhostAgent}{\emph{blue-ghost agent}}
\newcommand{\powerPillAgent}{\emph{power pill agent}}
\newcommand{\fearGhostAgent}{\emph{fear-ghosts agent}}

\newcommand{\taskone}{agent understanding task}
\newcommand{\tasktwo}{agent comparison task}

\newcommand{\taskonebig}{Agent Understanding Task}
\newcommand{\tasktwobig}{Agent Comparison Task}

\begin{document}
%%% The following commands remove the headers in your paper. For final 
%%% papers, these will be inserted during the pagination process.

\pagestyle{fancy}
\fancyhead{}

%%% The next command prints the information defined in the preamble.

\maketitle 

%%%%%%%%%%%%%%%%%%%%%%%%%%%%%%%%%%%%%%%%%%%%%%%%%%%%%%%%%%%%%%%%%%%%%%%%

\section{Introduction}
Modern Reinforcement Learning (RL) agents use increasingly complex state spaces and deep learning algorithms, making the decisions and strategies of such agents hard to understand \citep{Heuillet2021}.
At the same time, these deep RL agents are being deployed into increasingly high-risk domains like healthcare, autonomous driving, and robotic navigation \citep{yu2021,kiran2022,fan2020}.
In such domains, it is crucial to be able to understand the agents to enable appropriate use of them and to facilitate human-agent cooperation \citep{silva2022explainable}.
\begin{figure}
    \small
    \centering         
    \begin{minipage}{0.35\linewidth}
    \centering
    \includegraphics[width=0.9\linewidth]{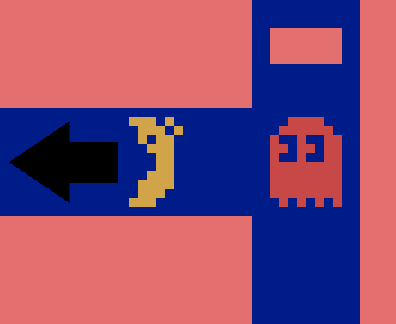}
    (a) Original state where Pacman goes left.
    \end{minipage}
    \hspace{0.1\linewidth}
    \begin{minipage}{0.35\linewidth}
    \centering
     \includegraphics[width=0.9\linewidth]{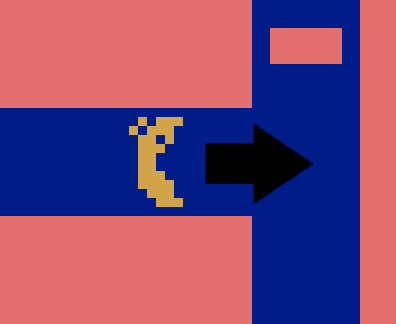}
    (b) Counterfactual state where Pacman goes right. 
    \end{minipage}
    \caption{
    Example for a counterfactual explanation: In the original situation (a), the agent does not take the fastest path to the pill in the top right corner.
It is unclear if the agent is afraid of the ghost or does not recognize the shortest path. 
The counterfactual state (b) shows that the agent would have taken the fastest path to the pill if the ghost was not there. 
This indicates that the agent is afraid of the ghost.
    }
    \label{fig:motivation_example}
    \Description{
    Situation (a) shows Pacman on the left of an intersection. On the right side of Pacman, in the intersection, there is a red ghost. Above that ghost there is pill. An arrow shows that Pacman will move to the left.
    Situation (b) shows the same situation, but the red ghost is not there. Here, an arrow shows that Pacman will move to the right.
    }
\end{figure}
One prominent paradigm to make the decisions of intelligent agents transparent and comprehensible are so-called \emph{Counterfactual Explanations}. 
By providing an alternative reality where the agent would have made a different decision, these explanations follow a rather human way of describing decisions \citep{miller19,byrne2019counterfactuals}.
For example, if a person would have to explain why a warehouse robot took a detour instead of directly moving to its desired target, they would probably give an explanation similar to \emph{If there was no production worker in the way, the robot would have moved straight to its target} - and, by doing so, give a counterfactual explanation of the warehouse robot's behavior.
Figure \ref{fig:motivation_example} shows a similar situation from the Atari game Pacman.

In other machine learning domains, such as image classification, 
counterfactual explanations are already frequently used. 
However, this is not the case for RL, as several factors make explaining the decisions of RL agents more challenging.
For one, RL agents are used for sequential decision-making tasks: their actions are not isolated.
These actions are part of a long-term strategy that might be influenced by delayed rewards.  
Secondly, RL agents are not trained on a given ground truth strategy. 
The reward function only indirectly specifies the agent's goals \citep{di2022goal}.
The emerging strategies might not be what humans would expect, even if the strategy is optimal for the reward function.
Finally, for RL agents, there is no direct counterpart to the training datasets used by supervised models.
Therefore, counterfactual explanation approaches for supervised models that utilize the training data cannot be applied to RL agents without adjustment \citep{wells2021}.

Due to the difficulties mentioned above, there is only one approach that focuses on creating counterfactual explanations for deep RL agents with visual input \citep{olson2021}.
This approach utilizes a complex combination of models where the final generator is only indirectly trained to change the action.
\citet{olson2021} show that their approach can be applied to a variety of RL environments and helps users identify a flawed agent.
With the help of their counterfactual explanations, users were able to differentiate between a normal RL agent for the Atari game Space Invaders and a flawed agent that did not see a specific in-game object.
For this task, it is sufficient for the counterfactual explanation to not change the particular object at all while other objects frequently change.
This clearly communicates that the unchanged object is irrelevant and ignored by the agent, implying that it is not seen at all.

But for counterfactual explanations to be employed more widely, they also have to be useful for more complex tasks.
According to \citet{hoffman2018metrics}, one of the main goals of a good explanation is to refine the user's mental model of the agent.
For RL agents, this includes understanding what strategy and intentions an agent pursues.
Another critical goal for explanations is that they should help users to calibrate their trust in different agents \citep{hoffman2018metrics}.
For RL agents, this entails that users should be able to choose fitting agents for specific problems, which is more complex than simply identifying defective agents.
The two aforementioned challenges require counterfactual explanations to not only convey \textit{what} objects need to change but also \textit{how} the objects need to be altered to change the agents' policy.

To tackle these challenges, this paper proposes a novel method for generating counterfactual explanations for RL agents with visual input. 
We do so by formulating the generation problem as a domain transfer problem where the domains are represented by sets of states that lead the agent to different actions.
Our approach is fully model-agnostic, easier to train than the approach presented by \citeauthor{olson2021}, and includes the counterfactual actions more directly into the training routine by solving an action-to-action domain transfer problem.
We evaluate our approach with computational metrics (e.g., how often do the counterfactuals change the agent's decision) and a user study using the Atari Learning Environment (ALE) \citep{bellemare2013ALE}, a common benchmark for RL agents with visual input.
In our user study, we present participants with different kinds of counterfactual explanations and investigate whether this helps them to understand the strategies of Pacman agents.
Furthermore, we investigate if the counterfactuals help them to calibrate their trust, so they can choose fitting agents for specific tasks (surviving or receiving points).

As such, the contributions of this paper are as follows:
We formulate a novel, model-agnostic approach for generating counterfactual explanations for RL agents. 
We demonstrate that our approach outperforms the previous method in several computational metrics.
Furthermore, we conduct a user study that shows, for the first time, that counterfactual explanations can help to understand the strategies of RL agents.
This user study also identifies current deficiencies of counterfactual explanations for RL agents that point the way for future work.

\section{Related Work}
Our work deals with post-hoc explanations that are generated for fully trained black-box agents. 
Recent years saw a plethora of work on such explanations for (deep) RL agents.
The literature often divides them by scope into global and local explanations.
Global explanations try to explain the agent's overall strategy.
This can be done by picking a subset of important state-action pairs that summarize the agent's strategy \citep{amir2018,huang2018} or by distilling the agent's policy into a simpler model like a finite state machine \citep{Danesh2021} or a soft decision tree \citep{coppens2019distilling}.
In this paper, we focus on local explanations that explain a specific decision of an agent.
The most common approach to local explanations for RL agents are Feature Attribution or Saliency Map methods \citep{zahavy2016graying,huber2019,puri2020}.
These methods try to identify the most important input features for a specific decision and highlight them, for example in a heatmap.
However, recent work questioned whether one can rely on post-hoc feature attribution to faithfully represent the agent's internal reasoning \citep{atrey2020,huber2022benchmarking}.
Furthermore, previous studies showed that saliency maps for visual RL agents are hard to understand for end-users \citep{huber2021LocalandGlobal}.
Counterfactual explanations are another type of local explanation.
Since they follow the human thinking paradigm of counterfactual reasoning, it is often argued that they are easier to interpret than feature attribution methods \citep{miller19,byrne2019counterfactuals}.

For classification models, there is a growing body of work on counterfactual explanations.
In 2017, Wachter et al. \cite{wachter2017} were the first to introduce counterfactual explanations into the XAI domain by defining them as an optimization problem.
Since then, various approaches to generate such counterfactuals were proposed, e.g., van \citet{DBLP:conf/pkdd/LooverenK21}, and \cite{DBLP:conf/icml/GoyalWEBPL19}.

As various research has observed that generating counterfactual explanations is, at its core, a generative problem, the use of generative models like Generative Adversarial Networks (GANs) quickly became prevailing in state-of-the-art counterfactual explanation generation algorithms.
E.g., Nemirovsky et al. \cite{DBLP:conf/uai/NemirovskyTXG22} proposed CounterGAN, a framework to build highly realistic and actionable counterfactual explanations.
Zhao et al. \cite{DBLP:conf/ijcai/ZhaoOK20} propose an approach for generating counterfactual image explanations by using text descriptions of relevant features of an image to be explained.
Furthermore, various specialized GAN-based algorithms were introduced to generate counterfactual explanations in the medical domain \cite{DBLP:journals/corr/abs-2110-14927, DBLP:journals/corr/abs-2101-04230, mertes2022ganterfactual}.
More recent frameworks for counterfactual explanation generation make use of the StyleGAN architecture, which implicitly models style-related aspects of an image, which makes it perfectly suitable for a whole range of image classification tasks \cite{DBLP:conf/iccv/LangGYWEHFIGIM21, DBLP:journals/corr/abs-2101-07563}.
As for a broad range of use cases, it is essential to be able to provide explanations for multiple counter-classes, various approaches have focused on that particular capability by using architectures based on StarGAN, an adversarial framework that was specifically designed for image translation between multiple domains \cite{zhao2020, DBLP:journals/corr/abs-2106-14556}.
One drawback of the aforementioned approaches for supervised learning is that their GANs are trained to transfer between domains given by the labeled classes from the classifier’s training dataset.
Then they add additional measures (e.g., loss functions \cite{mertes2022ganterfactual,zhao2020}), to ensure that the generated counterfactuals are actually classified as the desired class by the classifier that is to be explained.
This is not possible for RL agents that do not have a training set.
Furthermore, the additional measures are often not model-agnostic.

RL is often used to create counterfactual explanations for other models (for example in \citep{chen2021relace}).
However, to the best of the authors' knowledge, there is only one previous work on generating visual counterfactual explanations for RL agents \citep{olson2021}.
\citet{olson2021} train an encoder $E$ that creates an action-invariant latent representation of the agent's latent space. 
This is achieved by adversarially training $E$ in tandem with a discriminator $D$, where $D$ tries to predict the agent's action and $E$ aims to make the decision of $D$ as uncertain as possible.
In addition, they train a generative model $G$ to replicate states $s$ based on the action-invariant latent representation $E(s)$ and the agent's action probability distribution $\pi(s)$ for this state.
By providing $G$ with a counterfactual action distribution $\pi(s)'$, they obtain a state that is similar to $s$ but brings the agent's action distribution closer to the desired counterfactual distribution.
However, \citeauthor{olson2021} argue that an arbitrary counterfactual action distribution does not represent a realistic agent output and thus leads to unrealistic counterfactual states. To avoid this, they train an additional Wasserstein Auto Encoder and use it to perform gradient descent in the latent space of the agent towards an agent output that resembles the desired counterfactual action. 
\citeauthor{olson2021} refer to their approach as Counterfactual State Explanations (CSE), therefore we will also refer to it as \olson{} in this paper.

The \olson{} approach is fairly complex and requires extensive access to the agent's inner workings.
Furthermore, as \citeauthor{olson2021} mention themselves, the loss function of the generator $G$ does not directly force the resulting state $G(E(s),\pi(s)')$ to be classified as the counterfactual action distribution $\pi(s)'$. 
This is only learned indirectly by replicating states based on the action-invariant latent space and the desired action distribution $\pi(s)'$.
As we show in Section \ref{sec:eval_computational}, this does not seem to be enough to change the agent's decision correctly.
To solve those problems, we formulate a simpler counterfactual generation method that uses the counterfactual actions in a more direct way.

\section{Approach}
\label{sec:Approach}

\subsection{GANterfactual-RL}
\begin{figure}
    \centering
    \includegraphics[width=0.85\linewidth]{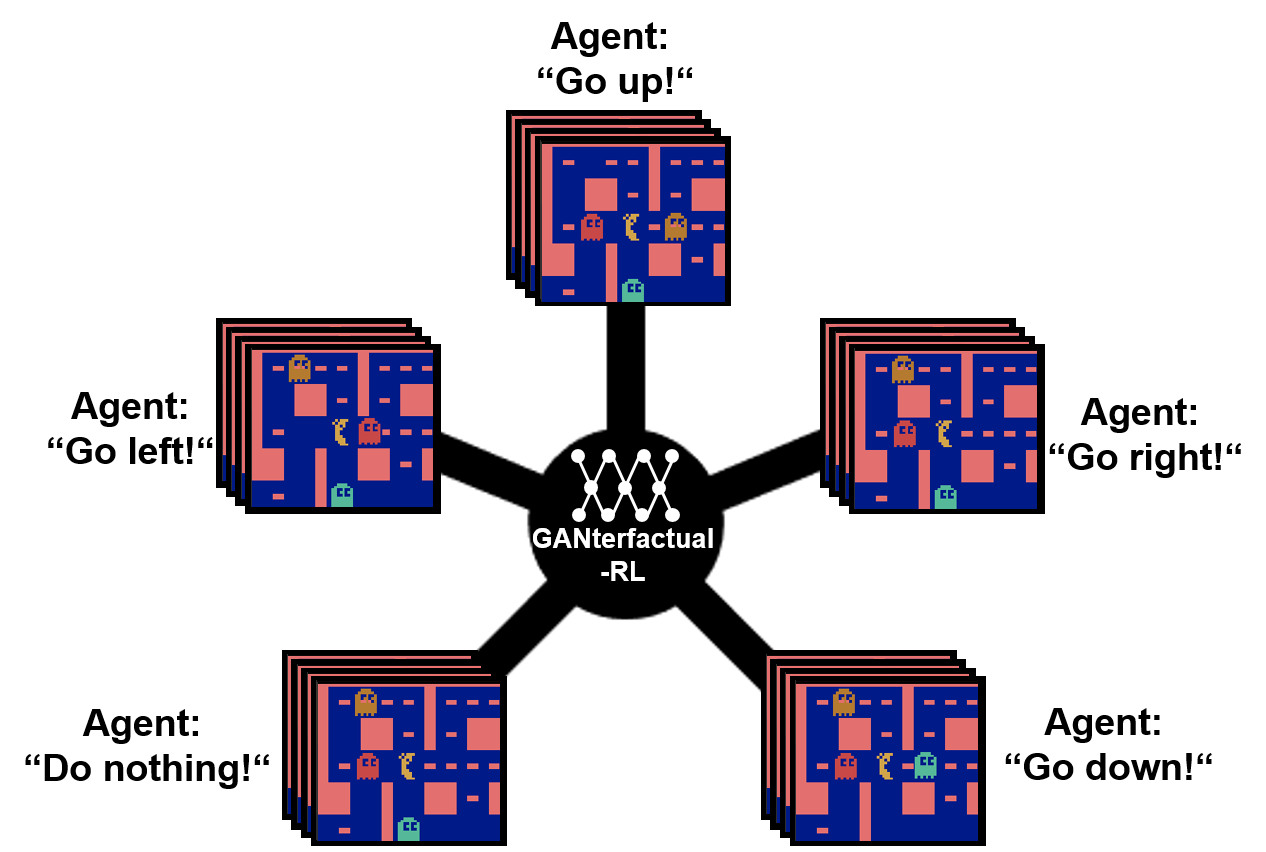}
    \caption{Schematic of our counterfactual generation approach.
    We formulate the problem as domain transfer where each domain represents an action.
    States are assigned to domains based on the action that the agent chooses for them.}
    \label{fig:counterfactuals_as_domain_transfer}
    \Description{
    In the center of the figure, there is a node that says "GANterfactual-RL" and shows a neural network icon. 
    From this center node, five black lines go to groups of images that show different states from the game Pacman. 
    The groups are respectively labeled with "Agent: "Go Left!"", "Agent: "Go Up!"", "Agent: "Go Right!"", "Agent: "Go Down!"", and "Agent: "Do nothing!"".
    }
\end{figure}

RL agents are usually employed in a Markov Decision Process (MDP) which consists of states $s\in S$, actions $a\in A$, and rewards $r$.
Given a state $s$, the goal of an RL agent $\pi:S \rightarrow A$ is to choose an action $\pi(s)$ that maximizes its cumulative future rewards. 
To explain such an agent, the objective of a counterfactual explanation approach for RL agents is defined as follows.
Given an original state $s$ and a desired counterfactual action $a'$, we want a counterfactual state $s'$ that makes the agent choose the desired action $\pi(s')=a'$.
Hereby, the original state $s$ should be altered as little as possible.
On an abstract level, the action $\pi(s)$ that the agent chooses for a state $s$ can be seen as a top-level feature that describes a combination of several underlying features which the agent considers to be relevant for its decision. 
Thus, the counterfactual state $s'$ should only change the features that are relevant to the agent's decision, while maintaining all other features not relevant to the decision.
This is similar to image-to-image translation, where features that are relevant for a certain image domain should be transformed into features leading to another image domain, while all other features have to be maintained (e.g., the background should remain constant when transforming horses to zebras).
Taken together, we can formulate the generation of counterfactual states for RL agents as a domain transfer problem similar to image-to-image translation:
The agent's action space $A$ defines the different domains $A_{i} = \{s \in S | \pi(s) = a_i\}$, where each state belongs to the domain that corresponds to the action that the agent chooses for this state (see Figure \ref{fig:counterfactuals_as_domain_transfer}).

To solve the reformulated domain transfer problem, we base our system on the StarGAN architecture \cite{choi2018stargan}, since RL agents usually use more than two actions.
The StarGAN architecture incorporates multiple loss components that can be reformulated to be applicable to the RL domain.
The first component, the so-called adversarial loss, leads the network to produce highly realistic states that look like states from the original environment.
Reformulated for the task of generating RL states, we define it as follows (following \citet{choi2018stargan} we use a Wasserstein objective with gradient penalty):
\begin{align*}
\mathcal{L}_{adv} &= {\mathbb{E}}_{s} \left[ {D}_{src}(s) \right] - {\mathbb{E}}_{s, a'}[{D}_{src}(G(s, a'))] \\
&- \lambda_{gp}{\mathbb{E}}_{\hat{s}} \left[(||\nabla_{\hat{s}} D_{src}(\hat{s})||_{2} - 1)^{2} \right],
\label{eq1}
\end{align*}
where ${D}_{src}$ is the StarGAN's \emph{discriminator} network and $G$ its \emph{generator} network.
The second loss component, which is specific to the StarGAN architecture, guides the generator network to produce states that lead to the desired counterfactual actions. It consists of two sub-objectives, one that is applied while the network is fed with original (\emph{real}) states from the training set (Eq. \ref{eq2}), and the other while the network is generating counterfactual states (Eq. \ref{eq3}):
\begin{equation}
\mathcal{L}_{cls}^{a} = {\mathbb{E}}_{s, a}[-\log{{D}_{cls}(a|s)}],
\label{eq2}
\end{equation}
\begin{equation}
\mathcal{L}_{cls}^{a'} ={\mathbb{E}}_{s, a'}[-\log{{D}_{cls}(a'|G(s, a'))}],
\label{eq3}
\end{equation}
where ${D}_{cls}$ refers to the StarGAN discriminator's classification output, which learns to approximate the action that the agent is performing in a particular state.
Further, as counterfactual states should be as close to the original states as possible, a \emph{Reconstruction Loss} is used.
This loss forces the network to only change features that are relevant to the agent's choice of action:
\begin{equation*}
\mathcal{L}_{rec} = {\mathbb{E}}_{s, a, a'} [{||s - G(G(s, a'), a)||}_{1} ]
\label{eq4}
\end{equation*}
Taken together, the whole loss of the StarGAN architecture, reformulated for RL counterfactual explanations, is defined as follows:
\begin{align*}
\mathcal{L}_{D} &=  - \mathcal {L}_{adv} +  {\lambda}_{cls}\thinspace\mathcal{L}_{cls}^{a}, \\
\mathcal{L}_{G} &=   \mathcal {L}_{adv} +  {\lambda}_{cls}\thinspace\mathcal{L}_{cls}^{a'} + 
{\lambda}_{rec}\thinspace\mathcal{L}_{rec},
\end{align*}
where ${\lambda}_{cls}$ and ${\lambda}_{rec}$ are weights controlling the corresponding loss component's relevance.
Since our approach utilizes a GAN architecture to generate counterfactuals for RL agents, we refer to it as \starGAN{}.

%%%%%%%%%%%%%%%%%%%%%%%%%%%%%%%%%%%%%%%% DATA SET
\subsection{Dataset Generation}
\label{sec:DatasetGeneration}

As described above, our \starGAN{} approach relies on training data in the form of state-action pairs.
\citet{olson2021} train their \olson{} approach on state-action pairs generated by concurrently running an MDP with a trained agent.
This strategy is simple but allows for little control over the training data, which can lead to the following complications:

\begin{itemize}
\item Frames extracted from a running MDP contain a temporal pattern since consecutive states typically have a high correlation. 
Such correlations and patterns can lead to bias and sub-optimal convergence during training.

\item For episodic MDPs, there is a high probability of reaching the same state throughout several episodes.
This is amplified by the fact that RL agents often learn to execute only a few optimal trajectories.
This results in duplicate samples that are effectively over-sampled during training.

\item RL agents generally do not execute each action equally frequently, since most environments contain actions that are useful more often than others.
This leads to an imbalanced amount of training samples per domain. 
\end{itemize}

To mitigate the aforementioned issues, we propose to generate datasets as follows:
Data is gathered by running a trained agent in an MDP.
Each state corresponds to one dataset sample and is labeled with the action that the agent chooses to execute in this state.
An $\epsilon$-greedy policy ($\epsilon {=} 0.2$ in our case) is used to increase the diversity of states reached over multiple episodes. 
State-action pairs with an explored (randomly chosen) action are not added to the dataset.
After the data is gathered, duplicates are removed. 
Then, a class balancing technique (under-sampling in our case) is used to account for over- or underrepresented actions.
Finally, the dataset is split into a training set, a test set, and potentially a validation set.

Most of these techniques are commonly used in other application domains of machine learning. However, to our best knowledge, this is the first work to generate and preprocess datasets for generating counterfactual explanations for RL agents.

\subsection{Application to Atari Domain}
\label{sec:agent_training}

\paragraph{Environment} 
The environments we use for our experiments are the Atari 2600 games MsPacman (henceforth referred to as Pacman) and Space Invaders, included in the Arcade Learning Environment (ALE) \cite{bellemare2013ALE}.
The ALE states are based on the raw pixel values of the game. 
Each input frame is cropped so that only the actual playing field remains. 
This removes components such as the score and life indicators which would allow participants to easily see which agent receives higher scores.
After that, we use the same preprocessing as \citet{mnih2015human}. 
Two steps from this preprocessing are particularly important for us.
First, the frames are gray-scaled and downsized.
Second, in addition to the current frame, the agent receives the last three preprocessed frames as input.
This allows the agent to detect temporal relations.
The ALE actions normally correspond to the meaningful actions achieved with an Atari 2600 controller (e.g. six actions for Space Invaders).
Since we wanted to use our Pacman agents in a user study we removed 4 redundant actions (e.g., \emph{Up \& Right}) whose effect differs between situations and is therefore hard to convey to participants.
This left us with 5 actions for Pacman (\emph{Do nothing, Up, Down, Left, Right}).

\emph{Agent Training.}
To evaluate participants' ability to differentiate between alternative agents and analyze their strategies, we modified the reward function of three Pacman agents.
This is a more natural method of obtaining different agents compared to withholding information from the agent as \citet{olson2021} did.
Furthermore, it results in agents that behave qualitatively differently.
Therefore participants have to actually analyze the agents' strategies instead of simply looking for objects that the agents ignore.
\begin{itemize}
	\setlength{\itemsep}{0pt}
    \item\textbf{\emph{Blue-Ghost Agent}}: This agent was trained using the default reward function of the ALE, where blue ghosts get the highest reward.
     \item \textbf{\emph{Power Pill Agent}}: This agent only received positive rewards for eating power pills.
    \item \textbf{\emph{Fear-Ghost Agent}}: This agent got a small positive reward of $1$ for every step in which it did not die to ghosts.
\end{itemize}
For training the first two Pacman agents, we use the DQN algorithm \cite{mnih2015human}.
Each agent was trained for 5 Million steps. 
The \fearGhostAgent{} was trained using the ACER algorithm \citep{wang2016sample} for 10M steps.
At the end of the training period, the best-performing policy is restored.
For all three agents, we build upon the OpenAI baselines \citep{baselines2017} repository.
For Space Invaders, we used the two Asynchronous Advantage Actor-Critic (A3C) agents trained by \citet{olson2021}.
For training details, we refer to their paper.
One agent is trained normally, while the other agent is flawed and does not see the laser cannon at the bottom of the screen.

\emph{\starGAN{} on Atari.}
To generate human-understandable counterfactual explanations for our Atari agents, the generated counterfactual states should represent the frames that humans see during gameplay.
That means we cannot train our \starGAN{} model on the preprocessed and stacked frames that the Atari agents use. 
Instead, we train it on the cropped RGB frames before preprocessing.
The only preprocessing we still use on those frames is a countermeasure against flickering objects in Atari games, which was proposed by \citet{mnih2015human}. 
While generating the dataset, we only save the most recent of the four stacked frames for each state $s$.
This frame generally influences the agent's decision the most.
For feeding the counterfactual frame back into the agent (e.g., to evaluate the approach), we stack it four times and then apply preprocessing.

Our implementation details can be found in the appendix. The full code is available online.\footnote{\url{https://github.com/hcmlab/GANterfactual-RL}}

\section{Computational Evaluation}
\label{sec:eval_computational}

\subsection{Used Metrics}

We evaluate our approach using the metrics \textit{validity} (or \textit{success rate}), \textit{proximity} (or \textit{cost}), \textit{sparsity}, and \textit{generation time}. We consider these metrics to be the most suitable and widely used metrics for image-based counterfactual explanations \citep{chen2021relace, Pawelczyk2021,Keane2021, Mothilal.01272020}. 

\textbf{Validity} captures the rate of CounterFactuals (CFs) that actually evoke the targeted action when fed to
the agent. With $N_T$ being true CFs (correctly changing the agent's action), $N_F$ being false CFs, and $N$  the total amount of evaluated CFs, this metric is defined as:
\begin{equation*}
    Validity = \frac{N_T}{N_T + N_F} = \frac{N_T}{N}
\end{equation*}

\textbf{Proximity} measures the similarity between an original state image and its CF via the $L1$-norm. We normalize the metric to measure the proximity in the range $[0, 1]$.
\begin{equation*}
    \label{eq:Proximity}
    Proximity(s, G) = 1 - \frac{1}{255 \cdot \mathcal{S}} ||s - G(s, a)||_1
\end{equation*}
where $s$ is the original state image, $G(s, a)$ is the generated CF for an arbitrary target action domain $a$ and $\mathcal{S}$ is the domain of color values of $s$ ($\mathcal{S}=3 \cdot Width \cdot Height$ for RGB-encoded images). 
The normalization with $255 \cdot \mathcal{S}$ assumes an 8-bit color encoding with color values in range $[0, 255]$. 
High proximity values are desirable since they indicate small adjustments to the original state.

\textbf{Sparsity} quantifies the number of unmodified pixel values between an original state image and its CF via the $L0$-norm (a pseudo-norm that counts the number of non-zero entries of a vector/matrix). The sparsity is normalized to the range $[0, 1]$ as well.
\begin{equation*}
    Sparsity(s, G) = 1 - \frac{1}{\mathcal{S}} ||s - G(s, a)||_0
\end{equation*}
A completely altered image has a sparsity of $0$, an unmodified image has a sparsity of $1$. High sparsity values are thus desirable.

\textbf{Generation Time} determines the time it takes to generate one CF with a trained generator, not including pre- or post-processing.

\subsection{Computational Results}

The computational results for the three Pacman agents are shown in Table \ref{tab:computational_eval_PacMan} and the results for the two Space Invaders agents in Table \ref{tab:computational_eval_SpaceInvaders}.
For the Pacman agents, we generated fully cleaned datasets (Section \ref{sec:DatasetGeneration}) and sampled 10\% of each action for the evaluation test set.
To show the contribution of our proposed dataset generation, we additionally trained a \starGAN{} model for the \blueGhostAgent{} without the steps proposed in Section \ref{sec:DatasetGeneration} and evaluated it on the test set from the clean dataset.
This dropped the validity to $0.45$ and sparsity to $0.50\pm 0.01$ while the other values stayed comparable.
To be more comparable to the results by \citet{olson2021}, we do not remove duplicates from the Space Invaders datasets and do not apply class balancing.
Here we create the test set by sampling 500 states for each action and removing all duplicates of these states from the training set.
Our \starGAN{} approach outperforms the \olson{} counterfactuals in every single metric.

\begin{table}[t]
  \caption{Computational evaluation results for the Pacman agents. \textit{Proximity}, \textit{sparsity} and \textit{generation time} are specified by \textit{mean $\pm$ standard deviation}.}
  \label{tab:computational_eval_PacMan}
    \small
  \begin{tabular}{lcccc}\toprule
    \textit{Approach} & \textit{Validity} ($\uparrow$) & \textit{Proximity} ($\uparrow$) & \textit{Sparsity} ($\uparrow$) & \textit{Gen. Time [s]} ($\downarrow$) \\ 
    \midrule
    \multicolumn{2}{l}{\emph{Blue-Ghost Agent}}\\
    Ours & 0.59 & 0.997 $\pm$ 0.001 & 0.73 $\pm$ 0.02 & 0.011 $\pm$ 0.012 \\
    \olson{} & 0.28 & 0.992 $\pm$ 0.002 & 0.33 $\pm$ 0.03 & 0.085 $\pm$ 0.021 \\ 
    \multicolumn{2}{l}{\emph{Power-Pill Agent}}\\
    Ours & 0.49 & 0.997 $\pm$ 0.001 & 0.70 $\pm$ 0.02 & 0.011 $\pm$ 0.008 \\
    \olson & 0.20 & 0.993 $\pm$ 0.002 & 0.32 $\pm$ 0.02 & 0.566 $\pm$ 0.731 \\ 
    \multicolumn{2}{l}{\emph{Fear-Ghost Agent}} \\
    Ours & 0.46 & 0.995 $\pm$ 0.001 & 0.45 $\pm$ 0.01 & 0.013 $\pm$ 0.014 \\
    \olson & 0.20 & 0.992 $\pm$ 0.002 & 0.32 $\pm$ 0.04 & 0.020 $\pm$ 0.017 \\ 
    \bottomrule
  \end{tabular}
\end{table}

\begin{table}[t]
  \caption{Computational evaluation results for the Space Invaders agents. \textit{Proximity}, \textit{sparsity} and \textit{generation time} are specified by \textit{mean $\pm$ standard deviation}.}
  \label{tab:computational_eval_SpaceInvaders}
    \small
  \begin{tabular}{lcccc}\toprule
    \textit{Approach} & \textit{Validity} ($\uparrow$) & \textit{Proximity} ($\uparrow$) & \textit{Sparsity} ($\uparrow$) & \textit{Gen. Time [s]} ($\downarrow$) \\
    \midrule
    \multicolumn{2}{l}{\emph{Normal Agent}} \\
    Ours & 0.70 & 0.998 $\pm$ 0.002 & 0.97 $\pm$ 0.02 & 0.011 $\pm$ 0.013 \\
    \olson & 0.18 & 0.995 $\pm$ 0.003 & 0.89 $\pm$ 0.05 & 6.180 $\pm$ 9.727 \\ 
    \multicolumn{2}{l}{\emph{Flawed Agent}} \\
    Ours & 0.53 & 0.998 $\pm$ 0.002 & 0.96 $\pm$ 0.01 & 0.011 $\pm$ 0.015 \\
    \olson & 0.17 & 0.995 $\pm$ 0.004 & 0.94 $\pm$ 0.01 & 0.020 $\pm$ 0.035 \\ 
    \bottomrule
  \end{tabular}
\end{table}

\begin{figure}
    \small
    \centering
    \newcommand{\myheight}{0.9 \baselineskip} % textbox height
    \newcommand{\firstColumn}{0.22 \linewidth} % textbox height
    \newcommand{\otherColumn}{0.25 \linewidth} % textbox height
    %First row
    \begin{minipage}[t]{\linewidth}
    \begin{minipage}[t]{0.22\linewidth}
    Agent:

    \

    Original Action:
    \end{minipage}
    \begin{minipage}[t]{0.25\linewidth}
    \centering
    \parbox[t][\myheight]{\linewidth}{\textbf{Pacman}}
    \parbox[t][\myheight]{\linewidth}{\textbf{Fear Ghosts}}
    \parbox[t][\myheight]{\linewidth}{Move Down}
    \end{minipage}
    \begin{minipage}[t]{0.25\linewidth}
    \centering
    \parbox[t][\myheight]{\linewidth}{\textbf{Space Invader}}
    \parbox[t][\myheight]{\linewidth}{\textbf{Flawed}}
    \parbox[t][\myheight]{\linewidth}{Right \& Fire}
    \end{minipage}
    \begin{minipage}[t]{0.25\linewidth}
    \centering
    \parbox[t][\myheight]{\linewidth}{\textbf{Space Invader}}
    \parbox[t][\myheight]{\linewidth}{\textbf{Normal}}
    \parbox[t][\myheight]{\linewidth}{Right \& Fire}
    \end{minipage}
    \end{minipage}
    % Second row
    \begin{minipage}{\linewidth}
        \begin{minipage}{\firstColumn}
            \vspace{0pt}
            Original State:
        \end{minipage}
        \begin{minipage}{0.25\linewidth}
        \vspace{0pt}
        \includegraphics[width=0.8\linewidth]{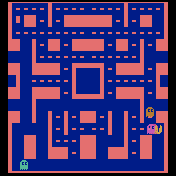}
        \end{minipage}
        \begin{minipage}{0.25\linewidth}
        \vspace{0pt}
        \includegraphics[width=0.8\linewidth]{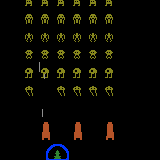}
        \end{minipage}
        \begin{minipage}{0.25\linewidth}
        \vspace{0pt}
        \includegraphics[width=0.8\linewidth]{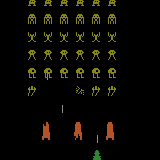}
        \end{minipage}
    \end{minipage}
     % Third row
    \begin{minipage}{\linewidth}
        \begin{minipage}{\firstColumn}
        Target Action:    
        \end{minipage}
        \begin{minipage}{\otherColumn}
        \parbox[c][\myheight]{\linewidth}{Move Up}    
        \end{minipage}
        \begin{minipage}{\otherColumn}
        \parbox[c][\myheight]{\linewidth}{Move Left}    
        \end{minipage}
        \begin{minipage}{\otherColumn}
        \parbox[c][\myheight]{\linewidth}{Move Left}
        \end{minipage}
    \end{minipage}
     % Fourth row
    \begin{minipage}{\linewidth}
        \begin{minipage}{\firstColumn}
        \vspace{0pt}
        CSE Counterfactual State:
        \end{minipage}
        \begin{minipage}{\otherColumn}
        \vspace{0pt}
        \includegraphics[width=0.8\linewidth]{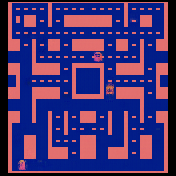}    
        \end{minipage}
        \begin{minipage}{\otherColumn}
        \vspace{0pt}
        \includegraphics[width=0.8\linewidth]{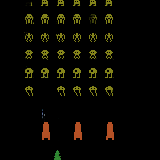}    
        \end{minipage}
        \begin{minipage}{\otherColumn}
         \vspace{0pt}
         \includegraphics[width=0.8\linewidth]{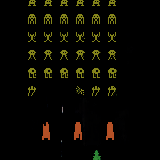}        
        \end{minipage}
    \end{minipage}
    % Fith row
    \begin{minipage}{\linewidth}
        \begin{minipage}{\firstColumn}
        \vspace{0pt}
        GANterfactual-RL Counterfactual State:
        \end{minipage}
        \begin{minipage}{\otherColumn}
        \vspace{0pt}
        \includegraphics[width=0.8\linewidth]{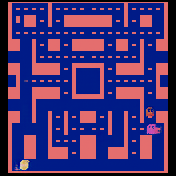}    
        \end{minipage}
        \begin{minipage}{\otherColumn}
        \vspace{0pt}
        \includegraphics[width=0.8\linewidth]{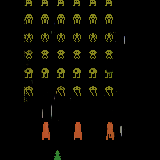}
        \end{minipage}
        \begin{minipage}{\otherColumn}
        \vspace{0pt}
        \includegraphics[width=0.8\linewidth]{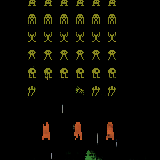}
        \end{minipage}
    \end{minipage}
    \caption{Example counterfactual states.
    Our approach does not change the \textit{Laser Cannon} (marked in blue) for the flawed agent, who does not see it, but changes it for the normal agent.
    }
    \label{fig:example_cfs}
    \Description{
    The figure shows three example counterfactual states.
    The first example is for the Pacman Fear Ghost agent. 
    Here, the original state shows Pacman on the bottom-right side of the labyrinth.
    Pacman is close to two ghosts and would die if it continues to move up.
    The original action is to move down, away from the ghosts. 
    The target action for the counterfactual states is "Move Up".
    The CSE counterfactual completely removes Pacman and the ghosts from the state.
    The GANterfactual-RL state moves Pacman to the bottom-left of the labyrinth where it can safely move up.
    The second example shows counterfactuals for the flawed Space Invader agent.
    The original action is "Right and Fire" and the counterfactual action is "Move Left".
    The CSE counterfactual removes all laser shots and one alien from the state. 
    The GANterfactual-RL counterfactual adds additional laser shots.
    None of the counterfactuals change the laser cannon.
    The last example shows counterfactuals for the normal Space Invader agent.
    The original action is "Right and Fire" and the counterfactual action is "Move Left".
    Here, both the CSE and the GANterfactual-RL counterfactual move the Laser Cannon slightly to the left.
    }
\end{figure}

Figure \ref{fig:example_cfs} shows example counterfactuals generated for the Pacman \fearGhostAgent{} and the two Space Invaders agents.
Additional examples for all our agents can be seen in the appendix.

\section{User Study}

\subsection{Study Design}

\subsubsection{Research Question and Hypothesis}
The research question for our study was which counterfactual explanations help users to understand the strategies of RL agents and help them to choose fitting agents for a specific task.
We hypothesized that our \starGAN{} method is more useful than the \olson{} method and is more useful than a presentation of the original states without counterfactuals.
Further, we thought that the counterfactuals generated by the \olson{} approach might mislead participants due to the low validity of the generated counterfactual explanations (see Section \ref{sec:eval_computational}).
Therefore, we hypothesized that only providing the original states is more useful than adding \olson{} counterfactuals.

\subsubsection{Dependent Variables and Main Tasks}

\paragraph{\taskonebig{}}
To measure whether participants understand the strategies of different agents and build a correct mental model of them, we used an \taskone{} inspired by \citet{hoffman2018metrics} and \citet{huber2021LocalandGlobal}.
Here, participants were presented with five states and the actions that the agent chooses in these states.
This was done for each of the three Pacman agents described in Section \ref{sec:agent_training} (one agent at a time).
The states were selected by the HIGHLIGHTS-Div algorithm \citep{amir2018}.
To this end, we let each trained agent play for additional $50$ episodes and chose the most important states according to HIGHLIGHTS-Div.
The resulting states show gameplay that is typical for the agent, without the need to manually select states that might be biased toward our approach.
Based on these states (and additional explanations depending on the condition), participants had to select up to two in-game objects that were most important for the agent's strategy from a list of objects (Pacman, normal pills, power pills, ghosts, blue ghosts, or cherries). 
As described in Section \ref{sec:agent_training}, each agent, strongly focuses on a different single in-game object depending on their reward function (e.g., the \fearGhostAgent{} focuses on normal ghosts).
If the participants select this object and none of the other objects, they receive a point.
The only exception is Pacman.
Every agent heavily relies on the position of Pacman as a source of information.
Therefore, participants receive the point whether they select Pacman or not.

\emph{\tasktwobig{}.}
To measure how well the participants' trust is calibrated, we used an \tasktwo{} inspired by \citet{amir2018} and \citet{miller2022trust}.
Here, we implicitly measure if the participants' trust is appropriate by asking them, for each possible pair of the three Pacman agents, which agent they would like to play on their behalf to obtain certain goals.
Since a single agent can be good for one goal but bad for another, this requires a deeper analysis than the distinction between a normal and a defective agent.
For each pair, the participants are shown their own descriptions of each agent from the \taskone{} and the same states and explanations that they saw during the \taskone{}.
Then they have to decide which agent should play on their behalf to achieve more points and which agent should play on their behalf to survive longer.
We know the ground truth for this by measuring the agents' average score and amount of steps for the $50$ episodes used to find the HIGHLIGHTS states.
The amount of steps that the \blueGhostAgent{} and the \powerPillAgent{} survive is so close that we do not include this specific comparison in the evaluation.

% explanation satisfaction
\emph{Explanation Satisfaction.}
To measure the participant's subjective satisfaction, we use statements adapted from the Explanation Satisfaction Scale by \citet{hoffman2018metrics}.
Participants have to rate their agreement with each statement on a 5-point Likert scale. 
Participants' final rating was averaged over all those ratings, reversing the rating of negative statements.
We do this once after the \taskone{} and once after the \tasktwo{} in case there are satisfaction differences between the tasks.

\subsubsection{Conditions and Explanation Presentation}
\begin{figure}
    \centering
    \includegraphics[width=0.5\linewidth]{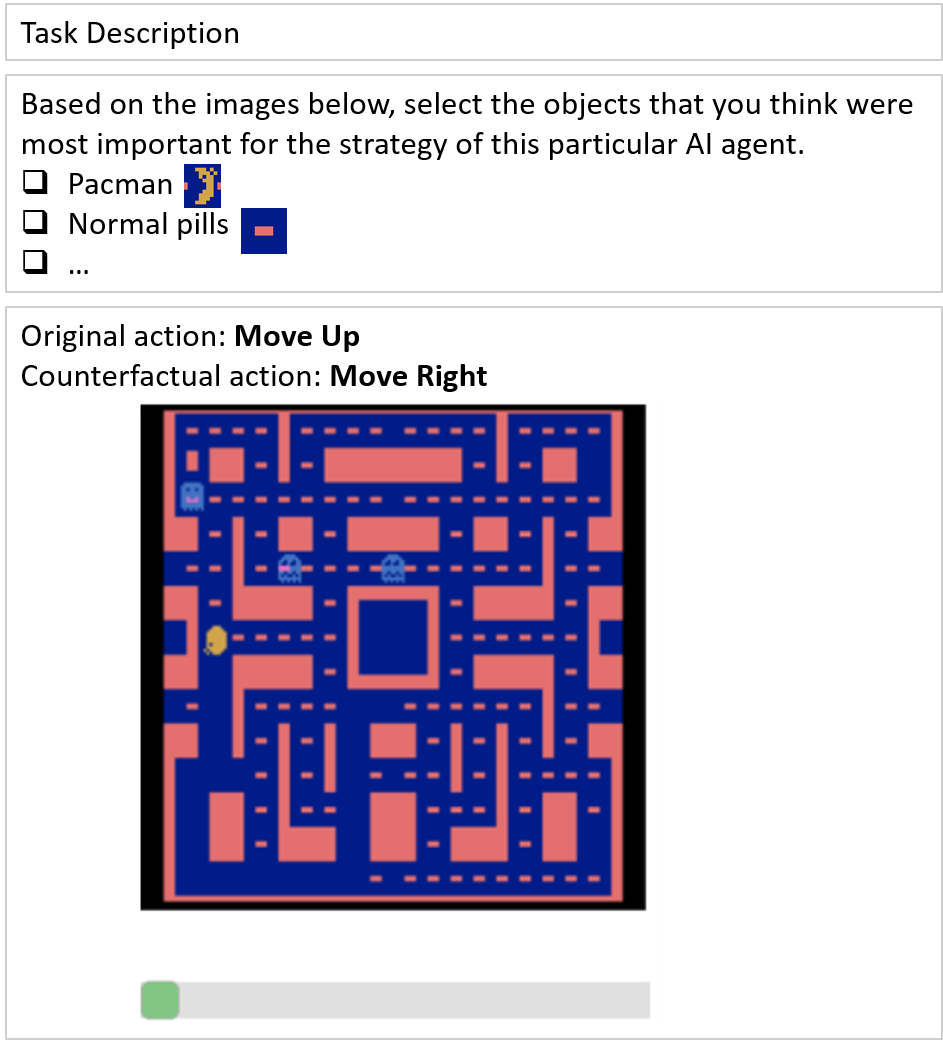}
    \caption{A simplified scheme of the beginning of our \taskone{} with a single example state.}
    \label{fig:study_scheme}
    \Description{
    The figure starts with a rectangle that symbolizes the description of the task that participants got.
    Below that, there is a rectangle that symbolizes the main question for the participants. 
    It reads, "Based on the images below, select the objects that you think were most important for the strategy of this particular AI agent." 
    The possible answers (e.g., "Pacman" or "Normal pills") are listed below the question as bullet points with the name and a picture of the object. 
    Below the main question there is a rectangle that shows an example state of the five states that participants saw for each agent.
    It says, "Original action: Move Up" and "Counterfactual action: Move right".
    Below this text, there is an image of a Pacman state and a slider that participants could use to morph the state into the counterfactual state.
    }
\end{figure}

We used three independent conditions, one \emph{\control{}} condition without explanations and two conditions where the states during the \taskone{} and the \tasktwo{} are accompanied by counterfactual explanations.
In the \emph{\olson{}} condition, the counterfactuals are generated by the approach from \citet{olson2021}, and in the \emph{\starGAN{}} condition the counterfactuals are generated by our proposed method.
The presentation of the counterfactual explanations is designed as follows.
For each state, we generate a single counterfactual state.
We were concerned that too many counterfactual states would cause too much cognitive load.
The way that MsPacman is implemented, actions that do nothing or move directly into a wall are ignored.
To generate meaningful counterfactual states, we limited the counterfactual action to turning around in a corridor and randomly selecting a new direction at an intersection (do not turn around).
The counterfactual states are presented by a slider under each state.
Moving the slider from left to right linearly interpolates the original state to the counterfactual state (\textit{per-pixel interpolation}).
The original and counterfactual actions are written above the state.
Figure \ref{fig:study_scheme} shows a simplified version of the beginning of our \taskone{}. 

\subsubsection{Procedure and Compensation}
After completing a consent form, participants were asked to answer demographic questions (age and gender) and questions regarding their experience with Pacman and their views on AI.
Then, they were shown a tutorial explaining the rules of the game Pacman and were asked to play the game to familiarize themselves with it.
To verify that participants understood the rules, they were asked to complete a quiz and were only allowed to proceed with the survey after answering all questions correctly.
Afterward, participants in the counterfactual conditions received additional information and another quiz regarding the counterfactual explanations.
Then, they proceeded to the \taskone{} which was repeated three times, once for each agent. 
The order of the agents was randomized.
After that, participants filled the explanation satisfaction scale and continued to the \tasktwo{}.
Again, this task was repeated three times, once for each possible agent pair, and the order was randomized.
Finally, participants had to complete another satisfaction scale for the \tasktwo{}.
Participants got a compensation of 5\$ for participating in the study.
As an incentive to do the tasks properly, they received a bonus payment of 10 cents for each point they get in the \taskone{} and 5 cents for each point in the \tasktwo{}.
The complete questionnaire can be seen in the appendix.
We preregistered our study online.\footnote{\url{https://aspredicted.org/m9fi5.pdf}}

\subsubsection{Participants}
We recruited participants through Amazon Mechanical Turk.
Participation was limited to Mechanical Turk Masters from the US, UK, or Canada (to ensure a sufficient English level) with a task approval rate greater than 95\% and without color vision impairment.
We conducted a power analysis with an estimated medium effect size of 0.7 based on previous similar experiments \citep{mertes2022ganterfactual,mertes2022alterfactual,huber2021LocalandGlobal}.
This determined that we need 28 participants per condition to achieve a power of 0.8. and a significance level of 0.05.
To account for participant exclusions, we recruited 30 participants per condition.
Participants were excluded if they did not look at any of the counterfactual explanations for any of the agents during the \taskone{}, if their textual answers were nonsensical or if they took considerably less time than the average.
This left us with $30$ participants in the \control{} condition, $28$ participants in the \olson{} condition, and $23$ in the \starGAN{} condition.

The distribution of age, AI experience, and Pacman experience was similar between the conditions (see the appendix). 
There was a difference in the gender distribution and the attitude towards AI between the conditions. 
The \control{} condition had 40\% female participants, the \olson{} condition had 32\% and the \starGAN{} condition had 26\%.
The mean attitude towards AI was the highest in the \starGAN{} condition and the lowest in the \control{} condition (see the appendix).

\subsection{Results}
\label{sec:results}

\begin{figure}
    \small
    \centering         
    \begin{minipage}{0.45\linewidth}
    \centering
    \includegraphics[width=\linewidth]{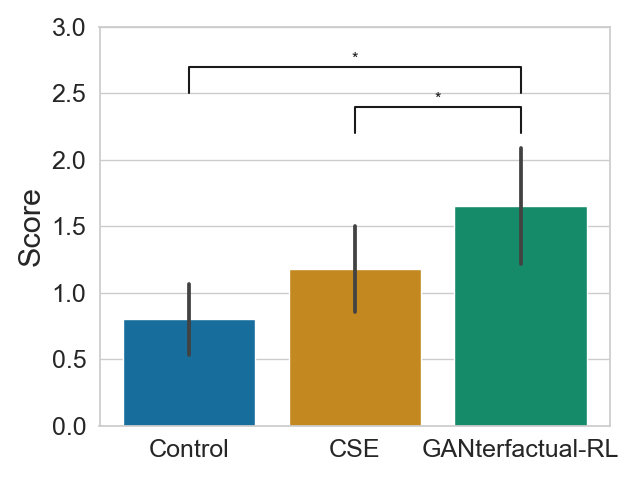}
    (a) Total score (summed over all three agents) for the \taskone{}.
    \end{minipage}
    \begin{minipage}{0.45\linewidth}
    \centering
    \includegraphics[width=\linewidth]{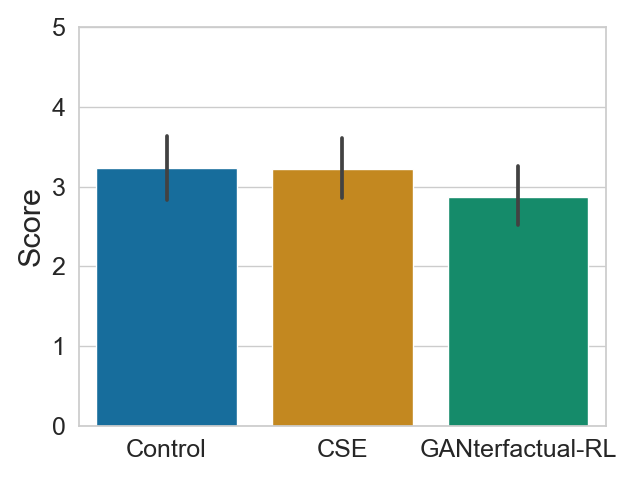}
    (b) Number of correct agent selections in the \tasktwo{} (Out of five).
    \end{minipage}
    \caption{Comparison of  participants' average performance in each task, by condition. Error bars show the 95\% CI.
    }
    \label{fig:total_score}
    \Description{
    Sub-figure (a) shows a bar plot for the results of the agent understanding task.
    For the control condition the mean was 0.8 and the confidence interval went from 0.54 to 1.06.
    For the CSE condition the mean was 1.18 and the confidence interval went from 0.86 to 1.50.
    For the GANterfactual-RL condition the mean was 1.65 and the confidence interval went from 1.23 to 2.07.
    The GANterfactual-RL condition was significantly higher then the CSE and Control conditions.
    Sub-figure (b) shows a bar plot for the results of the trust task.
    For the control condition the mean was 3.23 and the confidence interval went from 2.81 to 3.66.
    For the CSE condition the mean was 3.21 and the confidence interval went from 2.81 to 3.62.
    For the GANterfactual-RL condition the mean was 2.87 and the confidence interval went from 2.47 to 3.27.
    }
\end{figure}

\begin{figure}
    \centering
    \small
    \begin{minipage}{0.45\linewidth}
    \centering
    \includegraphics[width=\linewidth]{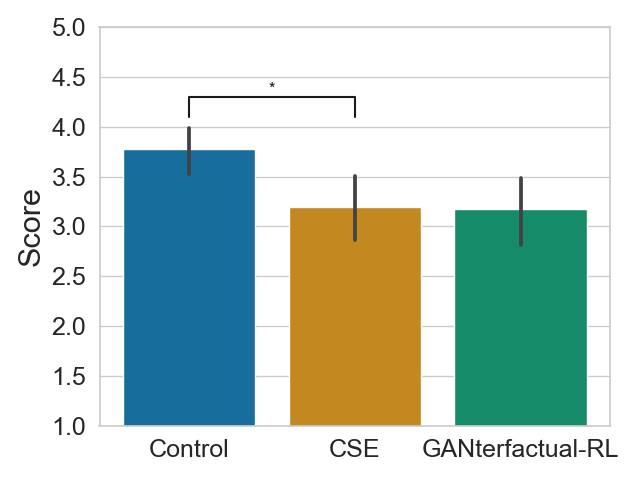}
    (a) \taskonebig{}.
    \end{minipage}
        \begin{minipage}{0.45\linewidth}
    \centering
    \includegraphics[width=\linewidth]{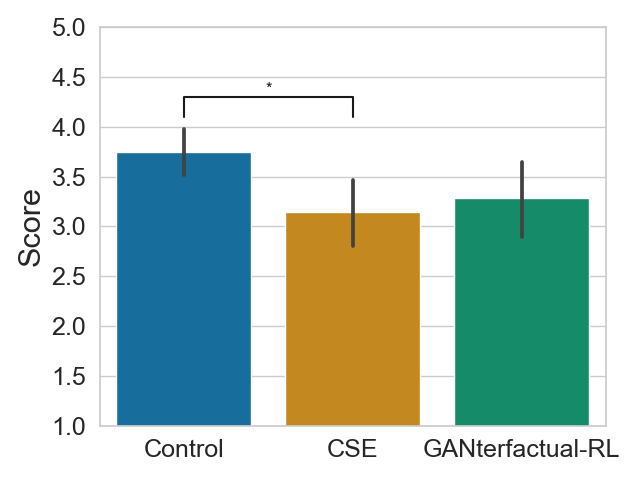}
    (b) \tasktwobig{}.
    \end{minipage}
    \caption{Comparison of participants' average explanation satisfaction in each task, by condition.}
    \label{fig:total_satisfaction}
    \Description{
    Sub-figure (a) shows a bar plot for the explanation satisfaction during agent understanding task.
    For the control condition the mean was 3.77 and the 95 percent confidence interval went from 3.54 to 4.01.
    For the CSE condition the mean was 3.20 and the confidence interval went from 2.88 to 3.51.
    For the GANterfactual-RL condition the mean was 3.17 and the confidence interval went from 2.82 to 3.53.
    The Control condition was significantly higher then the CSE condition.
    Sub-figure (b) shows a bar plot for the explanation satisfaction during the trust task.
    For the control condition the mean was 3.75 and the confidence interval went from 3.51 to 3.99.
    For the CSE condition the mean was 3.14 and the confidence interval went from 2.80 to 3.48.
    For the GANterfactual-RL condition the mean was 3.28 and the confidence interval went from 2.91 to 3.66.
    The Control condition was significantly higher then the CSE condition.
    }
\end{figure}

The results for the participants' scores during the main tasks can be seen in Figure \ref{fig:total_score}, while their explanation satisfaction values are shown in Figure \ref{fig:total_satisfaction}.
In the following, we will summarize the results of our main hypotheses, which we analyzed using non-parametric one-tailed Mann-Whitney U tests.

\newcommand{\sequal}{{=}}
\textbf{Counterfactuals helped participants to understand the agents' strategies.} 
In the \taskone{}, there was a significant difference between the \control{} condition ($M \sequal 0.8$) and the \starGAN{} condition ($M \sequal 1.65$), $U \sequal 181$, $p \sequal 0.001$, $r \sequal 0.477$.\footnote{$M$ is the mean and $r$ is rank biserial correlation.}  
Contrary to our hypothesis, the \control{} condition got lower scores than the \olson{} condition ($M \sequal 0.8$ vs $M \sequal 1.18$), $p \sequal 0.953$. 

\textbf{Our \starGAN{} explanations were significantly more useful than the \olson{} approach for understanding the agents' strategies.}
In the \taskone{}, the \olson{} condition got a mean score of $1.18$, while the \starGAN{} condition got a mean score of $1.65$ ($U \sequal 232$, $p \sequal 0.038$, $r \sequal 0.2795$).

\textbf{The increased understanding of the agents' strategies did not result in a more calibrated trust.} 
Contrary to our hypothesis, there were no significant differences in the trust task (\control{} vs \olson{}: $p \sequal 0.536$, \control{} vs. \starGAN{}: $p \sequal 0.852$, \olson{} vs \starGAN{}: $p \sequal 0.876$).

\textbf{Counterfactuals did not increase explanation satisfaction.}
Even though participants objectively had a better understanding of the agents' strategies, they did not feel more satisfied with them.
Participants in the \control{} condition were significantly more satisfied than participants in the \olson{} condition in both the \taskone{} (\control{}: $M \sequal 3.77$, \olson{}: $M \sequal 3.20$; $U \sequal249$, $p \sequal 0.004$, $r \sequal 0.4071$) and the \tasktwo{} (\control{}: $M \sequal 3.75$, \olson{}: $M \sequal 3.14$; $U \sequal 267$, $p \sequal 0.008$, $r \sequal 0.3643$).
Contrary to our expectations, the participants in the \starGAN{} condition were not more satisfied than the participants in the \control{} condition or the \olson{} condition in both the \taskone{} (\control{} vs. \starGAN{}: $p \sequal 0.996$, \olson{} vs \starGAN{}: $p \sequal 0.546$) or the \tasktwo{} (\control{} vs. \starGAN{}: $p \sequal 0.967$, \olson{} vs \starGAN{}: $p \sequal 0.334$).

\section{Discussion}

\subsection{Computational Evaluation}
Our computational evaluation shows that our proposed approach is correctly changing the agents' actions in 46\% to 70\% of the cases depending on the agent.
While this is not perfect, one has to consider that this is not a binary task but that the agents have 5 or 6 different actions.
Furthermore, CSE \cite{olson2021}, the only previous method that focuses on generating counterfactual explanations for RL agents, only successfully changed the agent's decision in 17\% to 28\% of the cases.
We can think of two reasons for the low validity values for the CSE approach.
First, they only incorporate the agent's action in their loss functions related to the latent space (where their discriminator and WAE were trained). 
The generation of the final pixels did not include constraints to faithfully ensure that a specific action was taken by the agent.
Second, their loss functions for the latent space focus on creating action-invariant states.
\citet{olson2021} showed that their \olson{} approach was useful for differentiating between a normal agent and a flawed agent. 
We think this is due to the fact that \olson{} is good at generating action-invariant states. 
This can help to identify the object that the flawed agent did not see since irrelevant objects are not changed for action-invariant states.
We found that our approach also does not change the irrelevant object for the flawed agent (illustrated in Figure \ref{fig:example_cfs}).
This demonstrates that the counterfactuals generated by our approach are similarly effective for identifying the flawed agent.
Looking at the distance between the original and the counterfactual states in pixel-space, we see that counterfactual states generated by our \starGAN{} approach on average have less distance to the original states and change fewer pixel values compared to the counterfactuals generated by the previous \olson{} method by \citet{olson2021}. 
This indicates that our \starGAN{} method is better at achieving the goal of finding the smallest possible modification of the original state to change the agent's decision. 
Since our method only requires a single forward pass to generate a counterfactual state, it is faster than the \olson{} method, which relies on potentially time-consuming gradient descent for the counterfactual generation.

\subsection{User Study}

Our user study showed that counterfactual explanations help users to understand which strategies different agents pursue.
In particular, our method was significantly more useful than both the \olson{} method and not providing counterfactuals.
Contrary to our hypothesis, even the counterfactuals generated by the \olson{} method resulted in a better understanding of the agents than not providing counterfactual explanations.
This demonstrates the usefulness of counterfactual explanations for RL agents even in more complex tasks than identifying defective agents.
Two recent studies evaluated the usefulness of other explanation techniques for understanding the strategies of RL agents in a similar way to our study.
\citet{huber2021LocalandGlobal} looked at saliency map explanations and found that they did not help more than showing HIGHLIGHTS states without saliency maps. 
Their participants achieved 37\% of the maximum possible score in their agent understanding task, while the participants with our counterfactual explanations obtained 50\%.
\citet{septon2022} investigated so-called reward decomposition explanations and found that they helped participants to achieve 60\% of the maximum score in their agent understanding task. 
However, reward decomposition is an intrinsic explanation method which the agent and the reward function have to be specifically designed for. 
Our counterfactual explanations resulted in only 10\% less average score even though they are post-hoc explanations that can be generated for already trained black-box agents.

Our \tasktwo{} showed that the increased understanding of the agent’s strategies through both counterfactual explanation methods did not help participants choose fitting agents for specific tasks.
For choosing the correct agent for a given problem, it is not enough to identify the strategies of the agents.
It also requires enough expertise in the environment (e.g., Pacman) to judge which strategy is better suited for the problem at hand.
For example, in Pacman, humans often assume that an agent that survives longer will accumulate more points in the long run. 
However, this is not necessarily the case since an aggressive agent can better exploit the very high rewards of eating blue ghosts.
Our results for this task are in line with the results of the \tasktwo{} for saliency maps by \citet{huber2021LocalandGlobal}.

Finally, our study showed that participants subjectively were not satisfied with the counterfactual explanations even though they objectively helped them to understand the agents. 
This might be due to the additional cognitive load of interpreting the explanations.
The two aforementioned studies \citep{huber2021LocalandGlobal,septon2022} also did not find a significant difference in user satisfaction for their local explanation techniques. 
Only the choice of states, which does not provide additional information, influenced the satisfaction in \citep{huber2021LocalandGlobal}.
However, our study is the first to see significantly higher satisfaction for the no-explanation condition than one of the two explanation conditions.
This indicates that counterfactuals are subjectively less satisfying than saliency maps or reward decomposition. 
One possible explanation for this is the visual quality of the counterfactuals.
Some participants from both counterfactual conditions commented that the counterfactuals had too many artifacts. 
One participant from the \starGAN{} condition for example wrote that \emph{"the counterfactuals were somewhat helpful, but they would have worked better if there were fewer or no artifacts"}.
Another possible reason for the low satisfaction is the presentation of the explanation.
Because our study primarily aimed at investigating the benefits and drawbacks of our specific counterfactual approach, we did not use a user-friendly explanatory system where different types of explanations are provided according to the requests of the explainee.

\section{Conclusion and Future Work}
In this work, we formulated a novel method for generating counterfactual explanations for RL agents.
This GANterfactual-RL method is fully model-agnostic,  which we demonstrate by applying it to three RL algorithms, two actor-critic methods, and one deep Q-learning method.
Using computational metrics, we show that our proposed method is better at correctly changing the agent's decision while modifying less of the original input and taking less time than the only previous method that focuses on generating visual counterfactuals for RL.
Furthermore, it significantly improved users' understanding of the strategies of different agents in a user study.

Our user study also identified two remaining deficiencies of counterfactual explanations. 
First, participants were subjectively not satisfied with the explanations, which might be due to unnatural artifacts in some counterfactuals.
Second, the counterfactuals did not help them to calibrate their trust in the agents. 
Future work should try to improve counterfactual explanations in these directions. 

While there is still room for improvement, we can confidently say that our approach can be considered the current state of the art for counterfactual explanations for RL agents with visual input.

%%% The following command should be issued somewhere in the first column 
%%% of the final page of your paper.

%%%%%%%%%%%%%%%%%%%%%%%%%%%%%%%%%%%%%%%%%%%%%%%%%%%%%%%%%%%%%%%%%%%%%%%%

%%% The acknowledgments section is defined using the "acks" environment
%%% (rather than an unnumbered section). The use of this environment 
%%% ensures the proper identification of the section in the article 
%%% metadata as well as the consistent spelling of the heading.

\begin{acks}
This paper was partially funded by the DFG through the Leibniz award of Elisabeth André (AN 559/10-1).
\end{acks}

%%%%%%%%%%%%%%%%%%%%%%%%%%%%%%%%%%%%%%%%%%%%%%%%%%%%%%%%%%%%%%%%%%%%%%%%

%%% The next two lines define, first, the bibliography style to be 
%%% applied, and, second, the bibliography file to be used.

\bibliographystyle{ACM-Reference-Format} 
\bibliography{main}


\begin{thebibliography}{47}

%%% the Web address.

\ifx \showCODEN    \undefined \def \showCODEN     #1{\unskip}     \fi
\ifx \showDOI      \undefined \def \showDOI       #1{#1}\fi
\ifx \showISBNx    \undefined \def \showISBNx     #1{\unskip}     \fi
\ifx \showISBNxiii \undefined \def \showISBNxiii  #1{\unskip}     \fi
\ifx \showISSN     \undefined \def \showISSN      #1{\unskip}     \fi
\ifx \showLCCN     \undefined \def \showLCCN      #1{\unskip}     \fi
\ifx \shownote     \undefined \def \shownote      #1{#1}          \fi
\ifx \showarticletitle \undefined \def \showarticletitle #1{#1}   \fi
\ifx \showURL      \undefined \def \showURL       {\relax}        \fi
% invisible to TeX
\providecommand\bibfield[2]{#2}
\providecommand\bibinfo[2]{#2}
\providecommand\natexlab[1]{#1}
\providecommand\showeprint[2][]{arXiv:#2}

\bibitem[\protect\citeauthoryear{Amir and Amir}{Amir and Amir}{2018}]%
        {amir2018}
\bibfield{author}{\bibinfo{person}{Dan Amir} {and} \bibinfo{person}{Ofra
  Amir}.} \bibinfo{year}{2018}\natexlab{}.
\newblock \showarticletitle{{HIGHLIGHTS:} Summarizing Agent Behavior to
  People}. In \bibinfo{booktitle}{\emph{Proceedings of the 17th International
  Conference on Autonomous Agents and MultiAgent Systems, {AAMAS} 2018,
  Stockholm, Sweden, July 10-15, 2018}}. \bibinfo{publisher}{International
  Foundation for Autonomous Agents and Multiagent Systems Richland, SC, {USA} /
  {ACM}}, \bibinfo{pages}{1168--1176}.
\newblock


\bibitem[\protect\citeauthoryear{Atrey, Clary, and Jensen}{Atrey
  et~al\mbox{.}}{2020}]%
        {atrey2020}
\bibfield{author}{\bibinfo{person}{Akanksha Atrey}, \bibinfo{person}{Kaleigh
  Clary}, {and} \bibinfo{person}{David~D. Jensen}.}
  \bibinfo{year}{2020}\natexlab{}.
\newblock \showarticletitle{Exploratory Not Explanatory: Counterfactual
  Analysis of Saliency Maps for Deep Reinforcement Learning}. In
  \bibinfo{booktitle}{\emph{8th International Conference on Learning
  Representations, {ICLR} 2020, Addis Ababa, Ethiopia, April 26-30, 2020}}.
\newblock
\urldef\tempurl%
\url{https://openreview.net/forum?id=rkl3m1BFDB}
\showURL{%
\tempurl}


\bibitem[\protect\citeauthoryear{Bellemare, Naddaf, Veness, and
  Bowling}{Bellemare et~al\mbox{.}}{2013}]%
        {bellemare2013ALE}
\bibfield{author}{\bibinfo{person}{Marc~G. Bellemare}, \bibinfo{person}{Yavar
  Naddaf}, \bibinfo{person}{Joel Veness}, {and} \bibinfo{person}{Michael
  Bowling}.} \bibinfo{year}{2013}\natexlab{}.
\newblock \showarticletitle{The Arcade Learning Environment: An Evaluation
  Platform for General Agents}.
\newblock \bibinfo{journal}{\emph{J. Artif. Intell. Res.}}
  \bibinfo{volume}{47} (\bibinfo{year}{2013}), \bibinfo{pages}{253--279}.
\newblock
\urldef\tempurl%
\url{https://doi.org/10.1613/jair.3912}
\showDOI{\tempurl}


\bibitem[\protect\citeauthoryear{Byrne}{Byrne}{2019}]%
        {byrne2019counterfactuals}
\bibfield{author}{\bibinfo{person}{Ruth M.~J. Byrne}.}
  \bibinfo{year}{2019}\natexlab{}.
\newblock \showarticletitle{Counterfactuals in Explainable Artificial
  Intelligence (XAI): Evidence from Human Reasoning}. In
  \bibinfo{booktitle}{\emph{Proceedings of the Twenty-Eighth International
  Joint Conference on Artificial Intelligence, {IJCAI-19}}}.
  \bibinfo{publisher}{International Joint Conferences on Artificial
  Intelligence Organization}, \bibinfo{pages}{6276--6282}.
\newblock
\urldef\tempurl%
\url{https://doi.org/10.24963/ijcai.2019/876}
\showDOI{\tempurl}


\bibitem[\protect\citeauthoryear{Chen, Silvestri, Tolomei, Zhu, Wang, and
  Ahn}{Chen et~al\mbox{.}}{2021}]%
        {chen2021relace}
\bibfield{author}{\bibinfo{person}{Ziheng Chen}, \bibinfo{person}{Fabrizio
  Silvestri}, \bibinfo{person}{Gabriele Tolomei}, \bibinfo{person}{He Zhu},
  \bibinfo{person}{Jia Wang}, {and} \bibinfo{person}{Hongshik Ahn}.}
  \bibinfo{year}{2021}\natexlab{}.
\newblock \showarticletitle{ReLACE: Reinforcement Learning Agent for
  Counterfactual Explanations of Arbitrary Predictive Models}.
\newblock \bibinfo{journal}{\emph{CoRR}}  \bibinfo{volume}{abs/2110.11960}
  (\bibinfo{year}{2021}).
\newblock


\bibitem[\protect\citeauthoryear{Choi, Choi, Kim, Ha, Kim, and Choo}{Choi
  et~al\mbox{.}}{2018}]%
        {choi2018stargan}
\bibfield{author}{\bibinfo{person}{Yunjey Choi}, \bibinfo{person}{Minje Choi},
  \bibinfo{person}{Munyoung Kim}, \bibinfo{person}{Jung-Woo Ha},
  \bibinfo{person}{Sunghun Kim}, {and} \bibinfo{person}{Jaegul Choo}.}
  \bibinfo{year}{2018}\natexlab{}.
\newblock \showarticletitle{Stargan: Unified generative adversarial networks
  for multi-domain image-to-image translation}. In
  \bibinfo{booktitle}{\emph{Proceedings of the IEEE conference on computer
  vision and pattern recognition}}. \bibinfo{pages}{8789--8797}.
\newblock


\bibitem[\protect\citeauthoryear{Coppens, Efthymiadis, Lenaerts, Now{\'e},
  Miller, Weber, and Magazzeni}{Coppens et~al\mbox{.}}{2019}]%
        {coppens2019distilling}
\bibfield{author}{\bibinfo{person}{Youri Coppens}, \bibinfo{person}{Kyriakos
  Efthymiadis}, \bibinfo{person}{Tom Lenaerts}, \bibinfo{person}{Ann Now{\'e}},
  \bibinfo{person}{Tim Miller}, \bibinfo{person}{Rosina Weber}, {and}
  \bibinfo{person}{Daniele Magazzeni}.} \bibinfo{year}{2019}\natexlab{}.
\newblock \showarticletitle{Distilling deep reinforcement learning policies in
  soft decision trees}. In \bibinfo{booktitle}{\emph{Proceedings of the IJCAI
  2019 workshop on explainable artificial intelligence}}.
  \bibinfo{pages}{1--6}.
\newblock


\bibitem[\protect\citeauthoryear{Danesh, Koul, Fern, and Khorram}{Danesh
  et~al\mbox{.}}{2021}]%
        {Danesh2021}
\bibfield{author}{\bibinfo{person}{Mohamad~H. Danesh}, \bibinfo{person}{Anurag
  Koul}, \bibinfo{person}{Alan Fern}, {and} \bibinfo{person}{Saeed Khorram}.}
  \bibinfo{year}{2021}\natexlab{}.
\newblock \showarticletitle{Re-understanding Finite-State Representations of
  Recurrent Policy Networks}. In \bibinfo{booktitle}{\emph{Proceedings of the
  38th International Conference on Machine Learning, {ICML} 2021, 18-24 July
  2021, Virtual Event}}. \bibinfo{pages}{2388--2397}.
\newblock
\urldef\tempurl%
\url{http://proceedings.mlr.press/v139/danesh21a.html}
\showURL{%
\tempurl}


\bibitem[\protect\citeauthoryear{Dhariwal, Hesse, Klimov, Nichol, Plappert,
  Radford, Schulman, Sidor, Wu, and Zhokhov}{Dhariwal et~al\mbox{.}}{2017}]%
        {baselines2017}
\bibfield{author}{\bibinfo{person}{Prafulla Dhariwal},
  \bibinfo{person}{Christopher Hesse}, \bibinfo{person}{Oleg Klimov},
  \bibinfo{person}{Alex Nichol}, \bibinfo{person}{Matthias Plappert},
  \bibinfo{person}{Alec Radford}, \bibinfo{person}{John Schulman},
  \bibinfo{person}{Szymon Sidor}, \bibinfo{person}{Yuhuai Wu}, {and}
  \bibinfo{person}{Peter Zhokhov}.} \bibinfo{year}{2017}\natexlab{}.
\newblock \bibinfo{title}{OpenAI Baselines}.
\newblock \bibinfo{howpublished}{\url{https://github.com/openai/baselines}}.
\newblock


\bibitem[\protect\citeauthoryear{Di~Langosco, Koch, Sharkey, Pfau, and
  Krueger}{Di~Langosco et~al\mbox{.}}{2022}]%
        {di2022goal}
\bibfield{author}{\bibinfo{person}{Lauro~Langosco Di~Langosco},
  \bibinfo{person}{Jack Koch}, \bibinfo{person}{Lee~D Sharkey},
  \bibinfo{person}{Jacob Pfau}, {and} \bibinfo{person}{David Krueger}.}
  \bibinfo{year}{2022}\natexlab{}.
\newblock \showarticletitle{Goal Misgeneralization in Deep Reinforcement
  Learning}. In \bibinfo{booktitle}{\emph{International Conference on Machine
  Learning}}. PMLR, \bibinfo{pages}{12004--12019}.
\newblock


\bibitem[\protect\citeauthoryear{Fan, Long, Liu, and Pan}{Fan
  et~al\mbox{.}}{2020}]%
        {fan2020}
\bibfield{author}{\bibinfo{person}{Tingxiang Fan}, \bibinfo{person}{Pinxin
  Long}, \bibinfo{person}{Wenxi Liu}, {and} \bibinfo{person}{Jia Pan}.}
  \bibinfo{year}{2020}\natexlab{}.
\newblock \showarticletitle{Distributed multi-robot collision avoidance via
  deep reinforcement learning for navigation in complex scenarios}.
\newblock \bibinfo{journal}{\emph{Int. J. Robotics Res.}} \bibinfo{volume}{39},
  \bibinfo{number}{7} (\bibinfo{year}{2020}).
\newblock
\urldef\tempurl%
\url{https://doi.org/10.1177/0278364920916531}
\showDOI{\tempurl}


\bibitem[\protect\citeauthoryear{Goyal, Wu, Ernst, Batra, Parikh, and
  Lee}{Goyal et~al\mbox{.}}{2019}]%
        {DBLP:conf/icml/GoyalWEBPL19}
\bibfield{author}{\bibinfo{person}{Yash Goyal}, \bibinfo{person}{Ziyan Wu},
  \bibinfo{person}{Jan Ernst}, \bibinfo{person}{Dhruv Batra},
  \bibinfo{person}{Devi Parikh}, {and} \bibinfo{person}{Stefan Lee}.}
  \bibinfo{year}{2019}\natexlab{}.
\newblock \showarticletitle{Counterfactual Visual Explanations}. In
  \bibinfo{booktitle}{\emph{Proceedings of the 36th International Conference on
  Machine Learning, {ICML} 2019, 9-15 June 2019, Long Beach, California,
  {USA}}} \emph{(\bibinfo{series}{Proceedings of Machine Learning Research},
  Vol.~\bibinfo{volume}{97})}, \bibfield{editor}{\bibinfo{person}{Kamalika
  Chaudhuri} {and} \bibinfo{person}{Ruslan Salakhutdinov}} (Eds.).
  \bibinfo{publisher}{{PMLR}}, \bibinfo{pages}{2376--2384}.
\newblock
\urldef\tempurl%
\url{http://proceedings.mlr.press/v97/goyal19a.html}
\showURL{%
\tempurl}


\bibitem[\protect\citeauthoryear{Heuillet, Couthouis, and
  Rodr{\'{\i}}guez}{Heuillet et~al\mbox{.}}{2021}]%
        {Heuillet2021}
\bibfield{author}{\bibinfo{person}{Alexandre Heuillet}, \bibinfo{person}{Fabien
  Couthouis}, {and} \bibinfo{person}{Natalia~D{\'{\i}}az Rodr{\'{\i}}guez}.}
  \bibinfo{year}{2021}\natexlab{}.
\newblock \showarticletitle{Explainability in deep reinforcement learning}.
\newblock \bibinfo{journal}{\emph{Knowl. Based Syst.}}  \bibinfo{volume}{214}
  (\bibinfo{year}{2021}), \bibinfo{pages}{106685}.
\newblock


\bibitem[\protect\citeauthoryear{Hoffman, Mueller, Klein, and Litman}{Hoffman
  et~al\mbox{.}}{2018}]%
        {hoffman2018metrics}
\bibfield{author}{\bibinfo{person}{Robert~R Hoffman}, \bibinfo{person}{Shane~T
  Mueller}, \bibinfo{person}{Gary Klein}, {and} \bibinfo{person}{Jordan
  Litman}.} \bibinfo{year}{2018}\natexlab{}.
\newblock \showarticletitle{Metrics for explainable {AI}: Challenges and
  prospects}.
\newblock \bibinfo{journal}{\emph{arXiv preprint arXiv:1812.04608}}
  (\bibinfo{year}{2018}).
\newblock


\bibitem[\protect\citeauthoryear{Huang, Bhatia, Abbeel, and Dragan}{Huang
  et~al\mbox{.}}{2018}]%
        {huang2018}
\bibfield{author}{\bibinfo{person}{Sandy~H. Huang}, \bibinfo{person}{Kush
  Bhatia}, \bibinfo{person}{Pieter Abbeel}, {and} \bibinfo{person}{Anca~D.
  Dragan}.} \bibinfo{year}{2018}\natexlab{}.
\newblock \showarticletitle{Establishing Appropriate Trust via Critical
  States}. In \bibinfo{booktitle}{\emph{2018 IEEE/RSJ International Conference
  on Intelligent Robots and Systems (IROS)}}. \bibinfo{pages}{3929--3936}.
\newblock
\urldef\tempurl%
\url{https://doi.org/10.1109/IROS.2018.8593649}
\showDOI{\tempurl}


\bibitem[\protect\citeauthoryear{Huber, Limmer, and André}{Huber
  et~al\mbox{.}}{2022}]%
        {huber2022benchmarking}
\bibfield{author}{\bibinfo{person}{Tobias Huber}, \bibinfo{person}{Benedikt
  Limmer}, {and} \bibinfo{person}{Elisabeth André}.}
  \bibinfo{year}{2022}\natexlab{}.
\newblock \showarticletitle{Benchmarking Perturbation-Based Saliency Maps for
  Explaining Atari Agents}.
\newblock \bibinfo{journal}{\emph{Frontiers in Artificial Intelligence}}
  \bibinfo{volume}{5} (\bibinfo{year}{2022}).
\newblock
\showISSN{2624-8212}
\urldef\tempurl%
\url{https://doi.org/10.3389/frai.2022.903875}
\showDOI{\tempurl}


\bibitem[\protect\citeauthoryear{Huber, Schiller, and Andr{\'e}}{Huber
  et~al\mbox{.}}{2019}]%
        {huber2019}
\bibfield{author}{\bibinfo{person}{Tobias Huber}, \bibinfo{person}{Dominik
  Schiller}, {and} \bibinfo{person}{Elisabeth Andr{\'e}}.}
  \bibinfo{year}{2019}\natexlab{}.
\newblock \showarticletitle{Enhancing Explainability of Deep Reinforcement
  Learning Through Selective Layer-Wise Relevance Propagation}. In
  \bibinfo{booktitle}{\emph{KI 2019: Advances in Artificial Intelligence}},
  \bibfield{editor}{\bibinfo{person}{Christoph Benzm{\"u}ller} {and}
  \bibinfo{person}{Heiner Stuckenschmidt}} (Eds.). \bibinfo{publisher}{Springer
  International Publishing}, \bibinfo{address}{Cham},
  \bibinfo{pages}{188--202}.
\newblock
\showISBNx{978-3-030-30179-8}


\bibitem[\protect\citeauthoryear{Huber, Weitz, Andr{\'{e}}, and Amir}{Huber
  et~al\mbox{.}}{2021}]%
        {huber2021LocalandGlobal}
\bibfield{author}{\bibinfo{person}{Tobias Huber}, \bibinfo{person}{Katharina
  Weitz}, \bibinfo{person}{Elisabeth Andr{\'{e}}}, {and} \bibinfo{person}{Ofra
  Amir}.} \bibinfo{year}{2021}\natexlab{}.
\newblock \showarticletitle{Local and global explanations of agent behavior:
  Integrating strategy summaries with saliency maps}.
\newblock \bibinfo{journal}{\emph{Artif. Intell.}}  \bibinfo{volume}{301}
  (\bibinfo{year}{2021}), \bibinfo{pages}{103571}.
\newblock
\urldef\tempurl%
\url{https://doi.org/10.1016/j.artint.2021.103571}
\showDOI{\tempurl}


\bibitem[\protect\citeauthoryear{Keane, Kenny, Delaney, and Smyth}{Keane
  et~al\mbox{.}}{2021}]%
        {Keane2021}
\bibfield{author}{\bibinfo{person}{Mark~T. Keane}, \bibinfo{person}{Eoin~M.
  Kenny}, \bibinfo{person}{Eoin Delaney}, {and} \bibinfo{person}{Barry Smyth}.}
  \bibinfo{year}{2021}\natexlab{}.
\newblock \showarticletitle{If Only We Had Better Counterfactual Explanations:
  Five Key Deficits to Rectify in the Evaluation of Counterfactual {XAI}
  Techniques}. In \bibinfo{booktitle}{\emph{Proceedings of the Thirtieth
  International Joint Conference on Artificial Intelligence, {IJCAI} 2021,
  Virtual Event / Montreal, Canada, 19-27 August 2021}}.
  \bibinfo{publisher}{ijcai.org}, \bibinfo{pages}{4466--4474}.
\newblock
\urldef\tempurl%
\url{https://doi.org/10.24963/ijcai.2021/609}
\showDOI{\tempurl}


\bibitem[\protect\citeauthoryear{Kiran, Sobh, Talpaert, Mannion, Sallab,
  Yogamani, and Pérez}{Kiran et~al\mbox{.}}{2022}]%
        {kiran2022}
\bibfield{author}{\bibinfo{person}{B~Ravi Kiran}, \bibinfo{person}{Ibrahim
  Sobh}, \bibinfo{person}{Victor Talpaert}, \bibinfo{person}{Patrick Mannion},
  \bibinfo{person}{Ahmad A.~Al Sallab}, \bibinfo{person}{Senthil Yogamani},
  {and} \bibinfo{person}{Patrick Pérez}.} \bibinfo{year}{2022}\natexlab{}.
\newblock \showarticletitle{Deep Reinforcement Learning for Autonomous Driving:
  A Survey}.
\newblock \bibinfo{journal}{\emph{IEEE Transactions on Intelligent
  Transportation Systems}} \bibinfo{volume}{23}, \bibinfo{number}{6}
  (\bibinfo{year}{2022}), \bibinfo{pages}{4909--4926}.
\newblock
\urldef\tempurl%
\url{https://doi.org/10.1109/TITS.2021.3054625}
\showDOI{\tempurl}


\bibitem[\protect\citeauthoryear{Lang, Gandelsman, Yarom, Wald, Elidan,
  Hassidim, Freeman, Isola, Globerson, Irani, and Mosseri}{Lang
  et~al\mbox{.}}{2021}]%
        {DBLP:conf/iccv/LangGYWEHFIGIM21}
\bibfield{author}{\bibinfo{person}{Oran Lang}, \bibinfo{person}{Yossi
  Gandelsman}, \bibinfo{person}{Michal Yarom}, \bibinfo{person}{Yoav Wald},
  \bibinfo{person}{Gal Elidan}, \bibinfo{person}{Avinatan Hassidim},
  \bibinfo{person}{William~T. Freeman}, \bibinfo{person}{Phillip Isola},
  \bibinfo{person}{Amir Globerson}, \bibinfo{person}{Michal Irani}, {and}
  \bibinfo{person}{Inbar Mosseri}.} \bibinfo{year}{2021}\natexlab{}.
\newblock \showarticletitle{Explaining in Style: Training a {GAN} to explain a
  classifier in StyleSpace}. In \bibinfo{booktitle}{\emph{2021 {IEEE/CVF}
  International Conference on Computer Vision, {ICCV} 2021, Montreal, QC,
  Canada, October 10-17, 2021}}. \bibinfo{publisher}{{IEEE}},
  \bibinfo{pages}{673--682}.
\newblock
\urldef\tempurl%
\url{https://doi.org/10.1109/ICCV48922.2021.00073}
\showDOI{\tempurl}


\bibitem[\protect\citeauthoryear{Looveren and Klaise}{Looveren and
  Klaise}{2021}]%
        {DBLP:conf/pkdd/LooverenK21}
\bibfield{author}{\bibinfo{person}{Arnaud~Van Looveren} {and}
  \bibinfo{person}{Janis Klaise}.} \bibinfo{year}{2021}\natexlab{}.
\newblock \showarticletitle{Interpretable Counterfactual Explanations Guided by
  Prototypes}. In \bibinfo{booktitle}{\emph{Machine Learning and Knowledge
  Discovery in Databases. Research Track - European Conference, {ECML} {PKDD}
  2021, Bilbao, Spain, September 13-17, 2021, Proceedings, Part {II}}}
  \emph{(\bibinfo{series}{Lecture Notes in Computer Science},
  Vol.~\bibinfo{volume}{12976})}, \bibfield{editor}{\bibinfo{person}{Nuria
  Oliver}, \bibinfo{person}{Fernando P{\'{e}}rez{-}Cruz},
  \bibinfo{person}{Stefan Kramer}, \bibinfo{person}{Jesse Read}, {and}
  \bibinfo{person}{Jos{\'{e}}~Antonio Lozano}} (Eds.).
  \bibinfo{publisher}{Springer}, \bibinfo{pages}{650--665}.
\newblock
\urldef\tempurl%
\url{https://doi.org/10.1007/978-3-030-86520-7 \_40}
\showDOI{\tempurl}


\bibitem[\protect\citeauthoryear{Matsui, Taki, Pham, Chikazoe, and
  Jimura}{Matsui et~al\mbox{.}}{2022}]%
        {DBLP:journals/corr/abs-2110-14927}
\bibfield{author}{\bibinfo{person}{Teppei Matsui}, \bibinfo{person}{Masato
  Taki}, \bibinfo{person}{Trung~Quang Pham}, \bibinfo{person}{Junichi
  Chikazoe}, {and} \bibinfo{person}{Koji Jimura}.}
  \bibinfo{year}{2022}\natexlab{}.
\newblock \showarticletitle{Counterfactual explanation of brain activity
  classifiers using image-to-image transfer by generative adversarial network}.
\newblock \bibinfo{journal}{\emph{Frontiers in Neuroinformatics}}
  \bibinfo{volume}{15} (\bibinfo{year}{2022}), \bibinfo{pages}{79}.
\newblock


\bibitem[\protect\citeauthoryear{Mertes, Huber, Weitz, Heimerl, and
  Andr{\'{e}}}{Mertes et~al\mbox{.}}{2022a}]%
        {mertes2022ganterfactual}
\bibfield{author}{\bibinfo{person}{Silvan Mertes}, \bibinfo{person}{Tobias
  Huber}, \bibinfo{person}{Katharina Weitz}, \bibinfo{person}{Alexander
  Heimerl}, {and} \bibinfo{person}{Elisabeth Andr{\'{e}}}.}
  \bibinfo{year}{2022}\natexlab{a}.
\newblock \showarticletitle{GANterfactual - Counterfactual Explanations for
  Medical Non-experts Using Generative Adversarial Learning}.
\newblock \bibinfo{journal}{\emph{Frontiers Artif. Intell.}}
  \bibinfo{volume}{5} (\bibinfo{year}{2022}), \bibinfo{pages}{825565}.
\newblock
\urldef\tempurl%
\url{https://doi.org/10.3389/frai.2022.825565}
\showDOI{\tempurl}


\bibitem[\protect\citeauthoryear{Mertes, Karle, Huber, Weitz, Schlagowski, and
  Andr{\'{e}}}{Mertes et~al\mbox{.}}{2022b}]%
        {mertes2022alterfactual}
\bibfield{author}{\bibinfo{person}{Silvan Mertes}, \bibinfo{person}{Christina
  Karle}, \bibinfo{person}{Tobias Huber}, \bibinfo{person}{Katharina Weitz},
  \bibinfo{person}{Ruben Schlagowski}, {and} \bibinfo{person}{Elisabeth
  Andr{\'{e}}}.} \bibinfo{year}{2022}\natexlab{b}.
\newblock \showarticletitle{Alterfactual Explanations - The Relevance of
  Irrelevance for Explaining {AI} Systems}.
\newblock \bibinfo{journal}{\emph{CoRR}}  \bibinfo{volume}{abs/2207.09374}
  (\bibinfo{year}{2022}).
\newblock
\urldef\tempurl%
\url{https://doi.org/10.48550/arXiv.2207.09374}
\showDOI{\tempurl}
\showeprint[arXiv]{2207.09374}


\bibitem[\protect\citeauthoryear{Miller}{Miller}{2019}]%
        {miller19}
\bibfield{author}{\bibinfo{person}{Tim Miller}.}
  \bibinfo{year}{2019}\natexlab{}.
\newblock \showarticletitle{Explanation in artificial intelligence: Insights
  from the social sciences}.
\newblock \bibinfo{journal}{\emph{Artif. Intell.}}  \bibinfo{volume}{267}
  (\bibinfo{year}{2019}), \bibinfo{pages}{1--38}.
\newblock
\urldef\tempurl%
\url{https://doi.org/10.1016/j.artint.2018.07.007}
\showDOI{\tempurl}


\bibitem[\protect\citeauthoryear{Miller}{Miller}{2022}]%
        {miller2022trust}
\bibfield{author}{\bibinfo{person}{Tim Miller}.}
  \bibinfo{year}{2022}\natexlab{}.
\newblock \showarticletitle{Are we measuring trust correctly in explainability,
  interpretability, and transparency research?}
\newblock \bibinfo{journal}{\emph{CoRR}}  \bibinfo{volume}{abs/2209.00651}
  (\bibinfo{year}{2022}).
\newblock
\urldef\tempurl%
\url{https://doi.org/10.48550/arXiv.2209.00651}
\showDOI{\tempurl}


\bibitem[\protect\citeauthoryear{Mnih, Kavukcuoglu, Silver, Rusu, Veness,
  Bellemare, Graves, Riedmiller, Fidjeland, Ostrovski, et~al\mbox{.}}{Mnih
  et~al\mbox{.}}{2015}]%
        {mnih2015human}
\bibfield{author}{\bibinfo{person}{Volodymyr Mnih}, \bibinfo{person}{Koray
  Kavukcuoglu}, \bibinfo{person}{David Silver}, \bibinfo{person}{Andrei~A
  Rusu}, \bibinfo{person}{Joel Veness}, \bibinfo{person}{Marc~G Bellemare},
  \bibinfo{person}{Alex Graves}, \bibinfo{person}{Martin Riedmiller},
  \bibinfo{person}{Andreas~K Fidjeland}, \bibinfo{person}{Georg Ostrovski},
  {et~al\mbox{.}}} \bibinfo{year}{2015}\natexlab{}.
\newblock \showarticletitle{Human-level control through deep reinforcement
  learning}.
\newblock \bibinfo{journal}{\emph{Nature}} \bibinfo{volume}{518},
  \bibinfo{number}{7540} (\bibinfo{year}{2015}), \bibinfo{pages}{529}.
\newblock


\bibitem[\protect\citeauthoryear{Mothilal, Sharma, and Tan}{Mothilal
  et~al\mbox{.}}{2020}]%
        {Mothilal.01272020}
\bibfield{author}{\bibinfo{person}{Ramaravind~Kommiya Mothilal},
  \bibinfo{person}{Amit Sharma}, {and} \bibinfo{person}{Chenhao Tan}.}
  \bibinfo{year}{2020}\natexlab{}.
\newblock \showarticletitle{Explaining machine learning classifiers through
  diverse counterfactual explanations}. In \bibinfo{booktitle}{\emph{FAT* '20:
  Conference on Fairness, Accountability, and Transparency, Barcelona, Spain,
  January 27-30, 2020}}. \bibinfo{publisher}{{ACM}}, \bibinfo{pages}{607--617}.
\newblock
\urldef\tempurl%
\url{https://doi.org/10.1145/3351095.3372850}
\showDOI{\tempurl}


\bibitem[\protect\citeauthoryear{Nemirovsky, Thiebaut, Xu, and
  Gupta}{Nemirovsky et~al\mbox{.}}{2022}]%
        {DBLP:conf/uai/NemirovskyTXG22}
\bibfield{author}{\bibinfo{person}{Daniel Nemirovsky}, \bibinfo{person}{Nicolas
  Thiebaut}, \bibinfo{person}{Ye Xu}, {and} \bibinfo{person}{Abhishek Gupta}.}
  \bibinfo{year}{2022}\natexlab{}.
\newblock \showarticletitle{CounteRGAN: Generating counterfactuals for
  real-time recourse and interpretability using residual GANs}. In
  \bibinfo{booktitle}{\emph{Uncertainty in Artificial Intelligence, Proceedings
  of the Thirty-Eighth Conference on Uncertainty in Artificial Intelligence,
  {UAI} 2022, 1-5 August 2022, Eindhoven, The Netherlands}}
  \emph{(\bibinfo{series}{Proceedings of Machine Learning Research},
  Vol.~\bibinfo{volume}{180})}, \bibfield{editor}{\bibinfo{person}{James
  Cussens} {and} \bibinfo{person}{Kun Zhang}} (Eds.).
  \bibinfo{publisher}{{PMLR}}, \bibinfo{pages}{1488--1497}.
\newblock
\urldef\tempurl%
\url{https://proceedings.mlr.press/v180/nemirovsky22a.html}
\showURL{%
\tempurl}


\bibitem[\protect\citeauthoryear{Olson, Khanna, Neal, Li, and Wong}{Olson
  et~al\mbox{.}}{2021}]%
        {olson2021}
\bibfield{author}{\bibinfo{person}{Matthew~L. Olson}, \bibinfo{person}{Roli
  Khanna}, \bibinfo{person}{Lawrence Neal}, \bibinfo{person}{Fuxin Li}, {and}
  \bibinfo{person}{Weng{-}Keen Wong}.} \bibinfo{year}{2021}\natexlab{}.
\newblock \showarticletitle{Counterfactual state explanations for reinforcement
  learning agents via generative deep learning}.
\newblock \bibinfo{journal}{\emph{Artif. Intell.}}  \bibinfo{volume}{295}
  (\bibinfo{year}{2021}), \bibinfo{pages}{103455}.
\newblock
\urldef\tempurl%
\url{https://doi.org/10.1016/j.artint.2021.103455}
\showDOI{\tempurl}


\bibitem[\protect\citeauthoryear{Pawelczyk, Bielawski, van~den Heuvel, Richter,
  and Kasneci}{Pawelczyk et~al\mbox{.}}{2021}]%
        {Pawelczyk2021}
\bibfield{author}{\bibinfo{person}{Martin Pawelczyk}, \bibinfo{person}{Sascha
  Bielawski}, \bibinfo{person}{Johannes van~den Heuvel},
  \bibinfo{person}{Tobias Richter}, {and} \bibinfo{person}{Gjergji Kasneci}.}
  \bibinfo{year}{2021}\natexlab{}.
\newblock \showarticletitle{{CARLA:} {A} Python Library to Benchmark
  Algorithmic Recourse and Counterfactual Explanation Algorithms}. In
  \bibinfo{booktitle}{\emph{Proceedings of the Neural Information Processing
  Systems Track on Datasets and Benchmarks 1, NeurIPS Datasets and Benchmarks
  2021, December 2021, virtual}}.
\newblock


\bibitem[\protect\citeauthoryear{Puri, Verma, Gupta, Kayastha, Deshmukh,
  Krishnamurthy, and Singh}{Puri et~al\mbox{.}}{2020}]%
        {puri2020}
\bibfield{author}{\bibinfo{person}{Nikaash Puri}, \bibinfo{person}{Sukriti
  Verma}, \bibinfo{person}{Piyush Gupta}, \bibinfo{person}{Dhruv Kayastha},
  \bibinfo{person}{Shripad Deshmukh}, \bibinfo{person}{Balaji Krishnamurthy},
  {and} \bibinfo{person}{Sameer Singh}.} \bibinfo{year}{2020}\natexlab{}.
\newblock \showarticletitle{Explain Your Move: Understanding Agent Actions
  Using Specific and Relevant Feature Attribution}. In
  \bibinfo{booktitle}{\emph{8th International Conference on Learning
  Representations, {ICLR}}}. \bibinfo{publisher}{OpenReview.net}.
\newblock


\bibitem[\protect\citeauthoryear{Russell and Norvig}{Russell and
  Norvig}{2016}]%
        {russell2016artificial}
\bibfield{author}{\bibinfo{person}{Stuart Russell} {and} \bibinfo{person}{Peter
  Norvig}.} \bibinfo{year}{2016}\natexlab{}.
\newblock \showarticletitle{Artificial Intelligence: A Modern Approach Global
  Edition}.
\newblock \bibinfo{journal}{\emph{Pearson}} (\bibinfo{year}{2016}).
\newblock


\bibitem[\protect\citeauthoryear{Schutte, Moindrot, H{\'{e}}rent, Schiratti,
  and J{\'{e}}gou}{Schutte et~al\mbox{.}}{2021}]%
        {DBLP:journals/corr/abs-2101-07563}
\bibfield{author}{\bibinfo{person}{Kathryn Schutte}, \bibinfo{person}{Olivier
  Moindrot}, \bibinfo{person}{Paul H{\'{e}}rent},
  \bibinfo{person}{Jean{-}Baptiste Schiratti}, {and} \bibinfo{person}{Simon
  J{\'{e}}gou}.} \bibinfo{year}{2021}\natexlab{}.
\newblock \showarticletitle{Using StyleGAN for Visual Interpretability of Deep
  Learning Models on Medical Images}.
\newblock \bibinfo{journal}{\emph{CoRR}}  \bibinfo{volume}{abs/2101.07563}
  (\bibinfo{year}{2021}).
\newblock


\bibitem[\protect\citeauthoryear{Septon, Huber, Andr{\'e}, and Amir}{Septon
  et~al\mbox{.}}{2022}]%
        {septon2022}
\bibfield{author}{\bibinfo{person}{Yael Septon}, \bibinfo{person}{Tobias
  Huber}, \bibinfo{person}{Elisabeth Andr{\'e}}, {and} \bibinfo{person}{Ofra
  Amir}.} \bibinfo{year}{2022}\natexlab{}.
\newblock \showarticletitle{Integrating Policy Summaries with Reward
  Decomposition for Explaining Reinforcement Learning Agents}.
\newblock \bibinfo{journal}{\emph{arXiv preprint arXiv:2210.11825}}
  (\bibinfo{year}{2022}).
\newblock


\bibitem[\protect\citeauthoryear{Silva, Schrum, Hedlund-Botti, Gopalan, and
  Gombolay}{Silva et~al\mbox{.}}{2022}]%
        {silva2022explainable}
\bibfield{author}{\bibinfo{person}{Andrew Silva}, \bibinfo{person}{Mariah
  Schrum}, \bibinfo{person}{Erin Hedlund-Botti}, \bibinfo{person}{Nakul
  Gopalan}, {and} \bibinfo{person}{Matthew Gombolay}.}
  \bibinfo{year}{2022}\natexlab{}.
\newblock \showarticletitle{Explainable Artificial Intelligence: Evaluating the
  Objective and Subjective Impacts of xAI on Human-Agent Interaction}.
\newblock \bibinfo{journal}{\emph{International Journal of Human--Computer
  Interaction}} (\bibinfo{year}{2022}), \bibinfo{pages}{1--15}.
\newblock


\bibitem[\protect\citeauthoryear{Singla, Eslami, Pollack, Wallace, and
  Batmanghelich}{Singla et~al\mbox{.}}{2023}]%
        {DBLP:journals/corr/abs-2101-04230}
\bibfield{author}{\bibinfo{person}{Sumedha Singla}, \bibinfo{person}{Motahhare
  Eslami}, \bibinfo{person}{Brian Pollack}, \bibinfo{person}{Stephen Wallace},
  {and} \bibinfo{person}{Kayhan Batmanghelich}.}
  \bibinfo{year}{2023}\natexlab{}.
\newblock \showarticletitle{Explaining the black-box smoothly - {A}
  counterfactual approach}.
\newblock \bibinfo{journal}{\emph{Medical Image Anal.}}  \bibinfo{volume}{84}
  (\bibinfo{year}{2023}), \bibinfo{pages}{102721}.
\newblock


\bibitem[\protect\citeauthoryear{Wachter, Mittelstadt, and Russell}{Wachter
  et~al\mbox{.}}{2017}]%
        {wachter2017}
\bibfield{author}{\bibinfo{person}{Sandra Wachter}, \bibinfo{person}{Brent
  Mittelstadt}, {and} \bibinfo{person}{Chris Russell}.}
  \bibinfo{year}{2017}\natexlab{}.
\newblock \showarticletitle{Counterfactual explanations without opening the
  black box: Automated decisions and the GDPR}.
\newblock \bibinfo{journal}{\emph{Harv. JL \& Tech.}}  \bibinfo{volume}{31}
  (\bibinfo{year}{2017}), \bibinfo{pages}{841}.
\newblock


\bibitem[\protect\citeauthoryear{Wang, Bapst, Heess, Mnih, Munos, Kavukcuoglu,
  and de~Freitas}{Wang et~al\mbox{.}}{2016}]%
        {wang2016sample}
\bibfield{author}{\bibinfo{person}{Ziyu Wang}, \bibinfo{person}{Victor Bapst},
  \bibinfo{person}{Nicolas Heess}, \bibinfo{person}{Volodymyr Mnih},
  \bibinfo{person}{Remi Munos}, \bibinfo{person}{Koray Kavukcuoglu}, {and}
  \bibinfo{person}{Nando de Freitas}.} \bibinfo{year}{2016}\natexlab{}.
\newblock \showarticletitle{Sample efficient actor-critic with experience
  replay}.
\newblock \bibinfo{journal}{\emph{arXiv preprint arXiv:1611.01224}}
  (\bibinfo{year}{2016}).
\newblock


\bibitem[\protect\citeauthoryear{Wells and Bednarz}{Wells and Bednarz}{2021}]%
        {wells2021}
\bibfield{author}{\bibinfo{person}{Lindsay Wells} {and} \bibinfo{person}{Tomasz
  Bednarz}.} \bibinfo{year}{2021}\natexlab{}.
\newblock \showarticletitle{Explainable {AI} and Reinforcement Learning - {A}
  Systematic Review of Current Approaches and Trends}.
\newblock \bibinfo{journal}{\emph{Frontiers Artif. Intell.}}
  \bibinfo{volume}{4} (\bibinfo{year}{2021}), \bibinfo{pages}{550030}.
\newblock
\urldef\tempurl%
\url{https://doi.org/10.3389/frai.2021.550030}
\showDOI{\tempurl}


\bibitem[\protect\citeauthoryear{White, Ngan, Phelan, Afgeh, Ryan,
  Reyes{-}Aldasoro, and d'Avila Garcez}{White et~al\mbox{.}}{2021}]%
        {DBLP:journals/corr/abs-2106-14556}
\bibfield{author}{\bibinfo{person}{Adam White}, \bibinfo{person}{Kwun~Ho Ngan},
  \bibinfo{person}{James Phelan}, \bibinfo{person}{Saman~Sadeghi Afgeh},
  \bibinfo{person}{Kevin Ryan}, \bibinfo{person}{Constantino~Carlos
  Reyes{-}Aldasoro}, {and} \bibinfo{person}{Artur d'Avila Garcez}.}
  \bibinfo{year}{2021}\natexlab{}.
\newblock \showarticletitle{Contrastive Counterfactual Visual Explanations With
  Overdetermination}.
\newblock \bibinfo{journal}{\emph{CoRR}}  \bibinfo{volume}{abs/2106.14556}
  (\bibinfo{year}{2021}).
\newblock
\showeprint[arXiv]{2106.14556}


\bibitem[\protect\citeauthoryear{Yu, Liu, Nemati, and Yin}{Yu
  et~al\mbox{.}}{2021}]%
        {yu2021}
\bibfield{author}{\bibinfo{person}{Chao Yu}, \bibinfo{person}{Jiming Liu},
  \bibinfo{person}{Shamim Nemati}, {and} \bibinfo{person}{Guosheng Yin}.}
  \bibinfo{year}{2021}\natexlab{}.
\newblock \showarticletitle{Reinforcement Learning in Healthcare: A Survey}.
\newblock  \bibinfo{volume}{55}, \bibinfo{number}{1}, Article
  \bibinfo{articleno}{5} (\bibinfo{year}{2021}), \bibinfo{numpages}{36}~pages.
\newblock
\showISSN{0360-0300}
\urldef\tempurl%
\url{https://doi.org/10.1145/3477600}
\showDOI{\tempurl}


\bibitem[\protect\citeauthoryear{Zahavy, Ben{-}Zrihem, and Mannor}{Zahavy
  et~al\mbox{.}}{2016}]%
        {zahavy2016graying}
\bibfield{author}{\bibinfo{person}{Tom Zahavy}, \bibinfo{person}{Nir
  Ben{-}Zrihem}, {and} \bibinfo{person}{Shie Mannor}.}
  \bibinfo{year}{2016}\natexlab{}.
\newblock \showarticletitle{Graying the black box: Understanding DQNs}. In
  \bibinfo{booktitle}{\emph{Proceedings of the 33nd International Conference on
  Machine Learning, {ICML} 2016, New York City, NY, USA, June 19-24, 2016}}.
  \bibinfo{pages}{1899--1908}.
\newblock
\urldef\tempurl%
\url{http://proceedings.mlr.press/v48/zahavy16.html}
\showURL{%
\tempurl}


\bibitem[\protect\citeauthoryear{Zhang and Dafoe}{Zhang and Dafoe}{2019}]%
        {zhang2019artificial}
\bibfield{author}{\bibinfo{person}{Baobao Zhang} {and} \bibinfo{person}{Allan
  Dafoe}.} \bibinfo{year}{2019}\natexlab{}.
\newblock \showarticletitle{Artificial intelligence: American attitudes and
  trends}.
\newblock \bibinfo{journal}{\emph{Available at SSRN 3312874}}
  (\bibinfo{year}{2019}).
\newblock


\bibitem[\protect\citeauthoryear{Zhao, Oyama, and Kurihara}{Zhao
  et~al\mbox{.}}{2020}]%
        {DBLP:conf/ijcai/ZhaoOK20}
\bibfield{author}{\bibinfo{person}{Wenqi Zhao}, \bibinfo{person}{Satoshi
  Oyama}, {and} \bibinfo{person}{Masahito Kurihara}.}
  \bibinfo{year}{2020}\natexlab{}.
\newblock \showarticletitle{Generating Natural Counterfactual Visual
  Explanations}. In \bibinfo{booktitle}{\emph{Proceedings of the Twenty-Ninth
  International Joint Conference on Artificial Intelligence, {IJCAI} 2020}},
  \bibfield{editor}{\bibinfo{person}{Christian Bessiere}} (Ed.).
  \bibinfo{publisher}{ijcai.org}, \bibinfo{pages}{5204--5205}.
\newblock
\urldef\tempurl%
\url{https://doi.org/10.24963/ijcai.2020/742}
\showDOI{\tempurl}


\bibitem[\protect\citeauthoryear{Zhao}{Zhao}{2020}]%
        {zhao2020}
\bibfield{author}{\bibinfo{person}{Yunxia Zhao}.}
  \bibinfo{year}{2020}\natexlab{}.
\newblock \showarticletitle{Fast Real-time Counterfactual Explanations}.
\newblock \bibinfo{journal}{\emph{CoRR}}  \bibinfo{volume}{abs/2007.05684}
  (\bibinfo{year}{2020}).
\newblock
\showeprint[arXiv]{2007.05684}


\end{thebibliography}

%%%%%%%%%%%%%%%%%%%%%%%%%%%%%%%%%%%%%%%%%%%%%%%%%%%%%%%%%%%%%%%%%%%%%%%%

\clearpage

\appendix

\section{Implementation Details}
\label{appendix:details}
In this section, we provide implementation details regarding the training of our counterfactual generation methods. 
Our full implementation can be found online.\footnote{\url{https://github.com/hcmlab/GANterfactual-RL}}

\subsection{Training Data}
For the size of our training data sets, we aimed for around $200 000$ states, since the StarGAN architecture from \citet{choi2018stargan} that we use in our approach was fine-tuned for the CelebA dataset, which contains around $200 000$ images.
To this end, we started by sampling $400 000$ states for each game and each RL agent.
For the Pacman agents, after duplicate removal and under-sampling (see Section 3.2), this leaves us with $230 450$ states for the \blueGhostAgent{},  $277 045$ states for the \powerPillAgent{} and $40 580$ states for the \fearGhostAgent{}.
For Space Invaders, the dataset size is only slightly reduced due to the removal of training samples that are duplicates of test samples.
For the normal agent, this resulted in $382 989$ states and for the flawed agent it resulted in $376 711$ states.

As is custom for the Atari environment \cite{bellemare2013ALE, mnih2015human}, we use a random amount (in the range $[0, 30]$) of initial \textit{Do Nothing} actions for each episode to make the games less deterministic.

\subsection{Training GANterfactual-RL}
For training the StarGAN within our GANterfactual-RL approach, we tried to stay as close to \citet{choi2018stargan} as possible.
We built our implementation upon the published source code\footnote{\url{https://github.com/yunjey/stargan}} of \citet{choi2018stargan}  and used their original settings.
The network architecture is the same as in \citet{choi2018stargan}. 
For the loss functions specified in the main paper, we use $\lambda_{cls}=1$, $\lambda_{rec}=10$, and $\lambda_{gp}=10$.
For training, we use an ADAM optimizer with $\alpha=0.0001$, $\beta_{1}=0.5$ and $\beta_{2}=0.999$.
The model is trained for $200,000$ batch iterations with a batch size of $16$.
The learning rate $\alpha$ linearly decays after half of the batch iterations are finished.
The Critic is updated 5 times per generator update during training.

One thing we change compared to \citet{choi2018stargan} is that we do not flip images horizontally during training.
This is an augmentation step that improves the generalization on datasets of face images. 
However, it is counterproductive for Atari frames since flipped frames would often leave the space of possible Atari states or change the action that the agent would select. 

\subsection{Training the Counterfactual State Explanations Model}
For training the counterfactual state explanation model proposed by \citet{olson2021}, we reuse their published source code\footnote{\url{https://github.com/mattolson93/counterfactual-state-explanations/}} to ensure comparability and reproducibility.
For this reason, we also use the same Training parameters and network architecture. 
The only change we had to make to the network architecture is that the size of the latent space of our DQN Pacman agents is $256$ compared to the Space Invaders agents in \citet{olson2021} that have a latent space size of $32$.

\section{User Study Demographics}
\label{appendix:demographics}
\begin{figure}[h]
    \centering
    \includegraphics[width=0.5\linewidth]{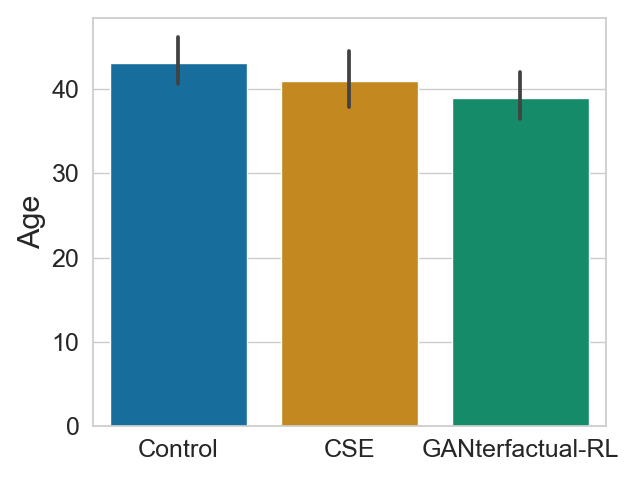}
    \caption{The participants' age per condition.}
    \label{fig:ages}
\end{figure}

In this section, we provide more details regarding the demographic of the participants in our user study.
As Fig. \ref{fig:ages} shows, the mean age for each condition was around 40.

\begin{figure}[h]
    \centering
     \includegraphics[width=0.8\linewidth]{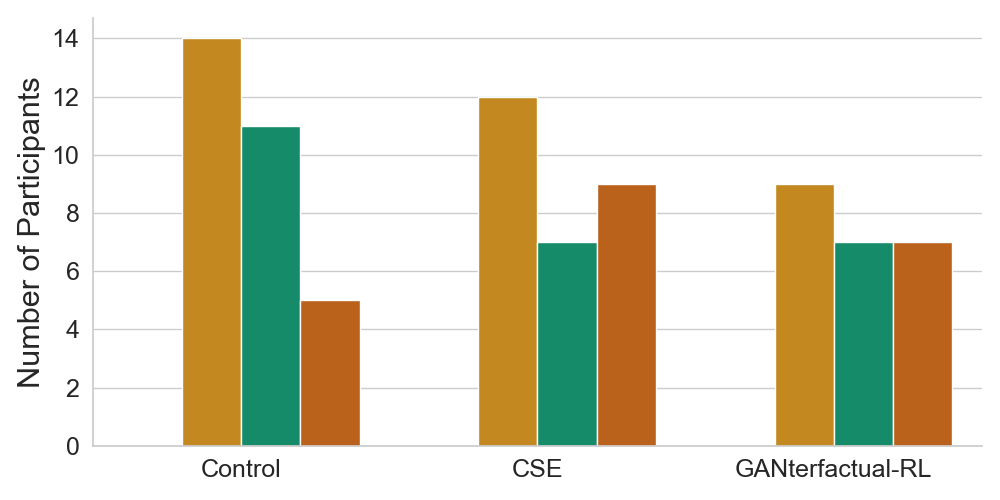}
    \caption{The Pacman experience across all conditions where the bars depict when the participants played Pacman the last time. From left to right the bars represent: ``more than 5 years ago'', ``less than 5 years ago'' and ``less than 1 year ago``. 
    }
    \label{fig:experience_Pamcan}
\end{figure}

We verified that participants in different conditions did not differ much in their AI experience and views and their Pacman experience.
To this end, we asked them when they played Pacman for the last time.
The results are shown in Figure \ref{fig:experience_Pamcan}.
    
\begin{figure}[!h]
    \centering
    \begin{minipage}{0.31\linewidth}
    \centering
    \includegraphics[width=\linewidth]{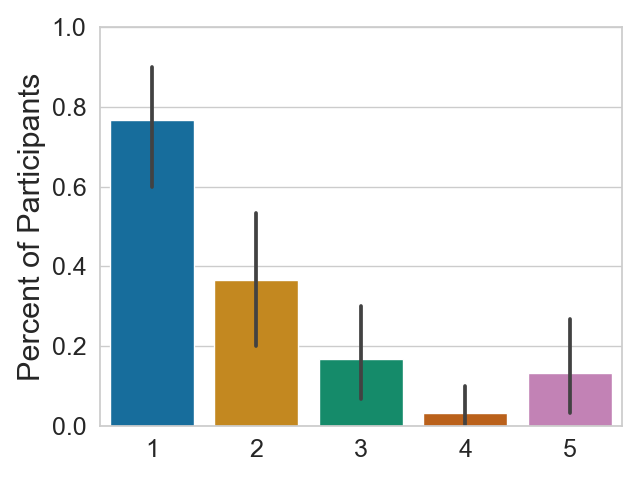}
    \control
    \end{minipage}
        \begin{minipage}{0.31\linewidth}
    \centering
    \includegraphics[width=\linewidth]{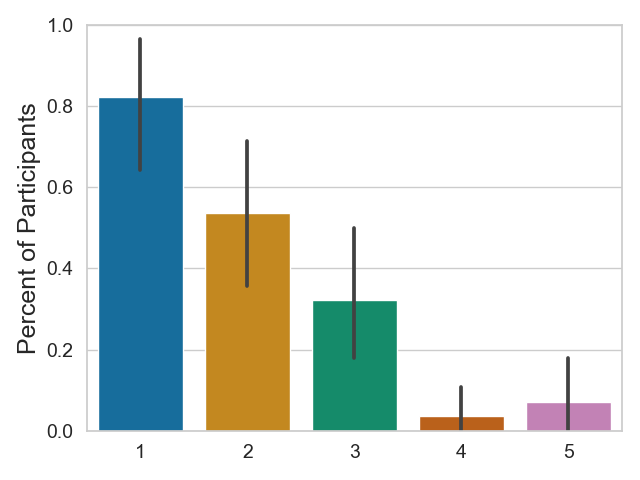}
    \olson
    \end{minipage}
    \begin{minipage}{0.31\linewidth}
    \centering
    \includegraphics[width=\linewidth]{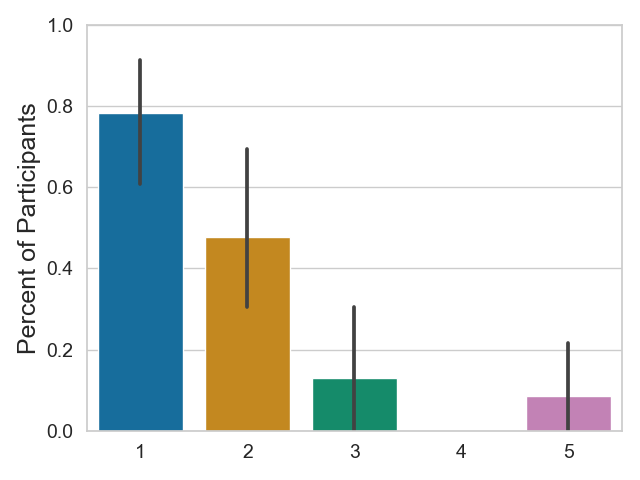}
    \starGAN
    \end{minipage}
    \caption{Distribution of the chosen AI experience items for each condition. The x-axis depicts the items described above.}
    \label{fig:experience_XAI}
\end{figure}

For the AI experience we adapted a description of AI from Zhang et al.~\cite{zhang2019artificial} and Russel \cite{russell2016artificial} to ``The following questions ask about Artificial Intelligence (AI). Colloquially, the term `artificial intelligence' is often used to describe machines (or computers) that mimic `cognitive' functions that humans associate with the human mind, such as `learning' and `problem solving'.'' 
After this description, participants had to select one or more of the following items:
\begin{itemize}
    \item 1: I know AI from the media.
    \item 2: I use AI technology in my private life.
    \item 3: I use AI technology in my work.
    \item 4: I took at least one AI-related course.
    \item 5: I do research on AI-related topics.
    \item Other:
\end{itemize}
The distribution of the items for each condition is shown in Fig. \ref{fig:experience_XAI}. The option \textit{Other} was never chosen.

\begin{figure}[!h]
    \centering
    \includegraphics[width=0.5\linewidth]{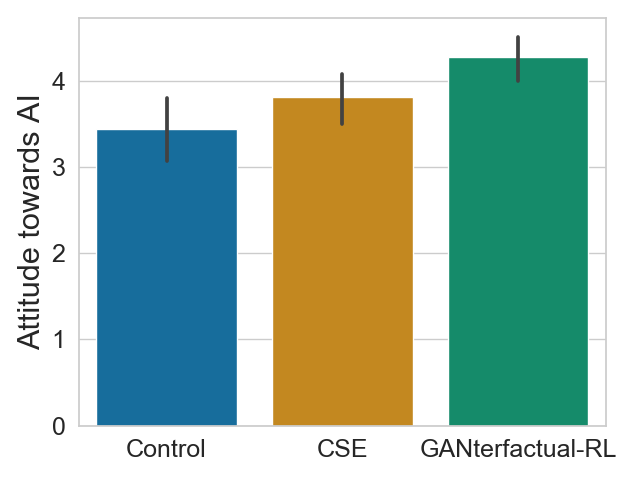}
    \caption{The average attitude towards AI, rated on a 5-point Likert scale. }
    \label{fig:attitude_AI}
\end{figure}

To measure the participants' attitude towards AI we adapted a question from Zhang et al~\cite{zhang2019artificial} and asked them to rate their answer to the question ``Suppose that AI agents would achieve high-level performance in more areas one day. How positive or negative do you expect the overall impact of such AI agents to be on humanity in the long run?'' on a 5-point Likert scale from ``Extremely negative'' to ``Extremely positive''.
The results are shown in Fig.~\ref{fig:attitude_AI}.

\section{Supplementary Results}
\label{appendix:results}
In this section, we present additional information about the results of the user study that did not fit the scope of the main paper.

\begin{figure}[h]
    \centering
    \includegraphics[width=0.6\linewidth]{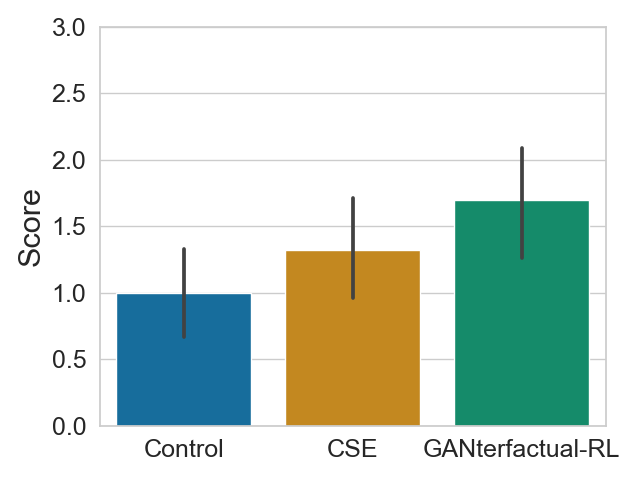}
    \caption{How often the participants in each condition selected Pacman as important for the agent's strategy.}
    \label{fig:selected_pacman}
\end{figure}

During our \taskone{} participants got the point for each agent independent of whether they selected Pacman as important since Pacman is the main source of information for all our agents.
We still wanted to exploratively look at how often the participants picked Pacman in each condition.
Figure \ref{fig:selected_pacman} shows that the results for picking Pacman are similar to the results for the participants' scores in this task.

\begin{figure}[h]
    \centering
    \includegraphics[width=0.6\linewidth]{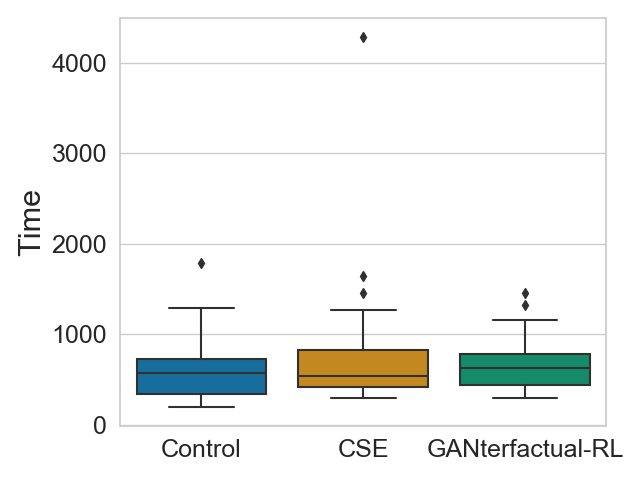}
    \caption{How much time (in seconds) the participants spent on our two main tasks.}
    \label{fig:times}
\end{figure}

Another value we exploratively looked at is how long the participants in each condition spent on doing the two main tasks (excluding the time they spent on the instructions, quizzes, and satisfaction questions).
Figure \ref{fig:times} shows that the time that participants spend on the main tasks does not differ much between conditions.

Finally, we want to report the average in-game score and survival time of our Pacman agents since we used this as ground truth for the \tasktwo{}.
The \blueGhostAgent{} got a mean score of $2035.6$ and survived for $708.36$ steps on average.
The \powerPillAgent{} got a mean score of $1488$ and survived for $696.4$ steps on average.
The \fearGhostAgent{} got a mean score of $944.4$ and survived for $6490.16$ steps on average.

\section{Example Counterfactuals}

\begin{figure*}
    \small
    \centering
    \newcommand{\myheight}{0.9 \baselineskip} % textbox height
    % state 0
    \begin{minipage}{0.19\linewidth}
    \centering
    \parbox[t][\myheight]{\linewidth}{Move Right}
    \includegraphics[width=\linewidth]{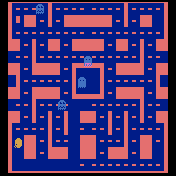}
    \parbox[t][\myheight]{\linewidth}{Move Down}
    \includegraphics[width=\linewidth]{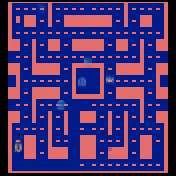}
    \includegraphics[width=\linewidth]{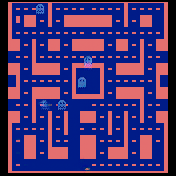}
    \end{minipage}
    %state 1
      \begin{minipage}{0.19\linewidth}
    \centering
    \parbox[t][\myheight]{\linewidth}{Move Up}
    \includegraphics[width=\linewidth]{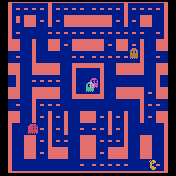}
    \parbox[t][\myheight]{\linewidth}{Move Left}
    \includegraphics[width=\linewidth]{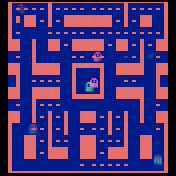}
    \includegraphics[width=\linewidth]{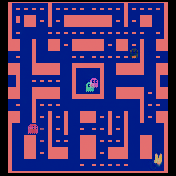}
    \end{minipage}
    % state 2
      \begin{minipage}{0.19\linewidth}
    \centering
    \parbox[t][\myheight]{\linewidth}{Move Up}
    \includegraphics[width=\linewidth]{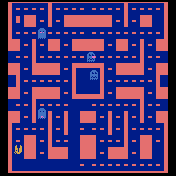}
    \parbox[t][\myheight]{\linewidth}{Move Down}
    \includegraphics[width=\linewidth]{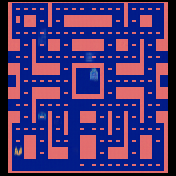}
    \includegraphics[width=\linewidth]{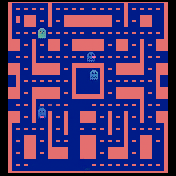}
    \end{minipage}
    % state 3
      \begin{minipage}{0.19\linewidth}
    \centering
    \parbox[t][\myheight]{\linewidth}{Move Down}
    \includegraphics[width=\linewidth]{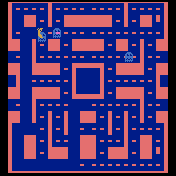}
    \parbox[t][\myheight]{\linewidth}{Move Right}
    \includegraphics[width=\linewidth]{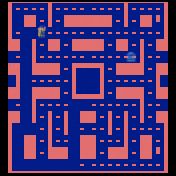}
    \includegraphics[width=\linewidth]{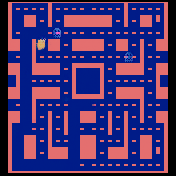}
    \end{minipage}
    % state 4
      \begin{minipage}{0.19\linewidth}
    \centering
    \parbox[t][\myheight]{\linewidth}{Move Up}
    \includegraphics[width=\linewidth]{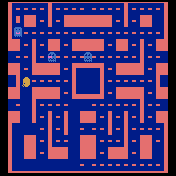}
    \parbox[t][\myheight]{\linewidth}{Move Right}
    \includegraphics[width=\linewidth]{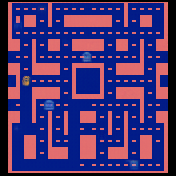}
    \includegraphics[width=\linewidth]{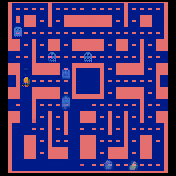}
    \end{minipage}
    \caption{Example counterfactual states for the \blueGhostAgent{}.
    The first row shows the original states and the second and third rows show counterfactual states by \citet{olson2021} and our \starGAN{} approach respectively. 
    The states and actions are the same states that were used during our user study and were chosen by the HIGHLIGHTS-Div algorithm \cite{amir2018}.
    }
    \label{fig:blue_ghost_examples}
\end{figure*}

\begin{figure*}
    \small
    \centering
    \newcommand{\myheight}{0.9 \baselineskip} % textbox height
    % state 0
    \begin{minipage}{0.19\linewidth}
    \centering
    \parbox[t][\myheight]{\linewidth}{Move Left}
    \includegraphics[width=\linewidth]{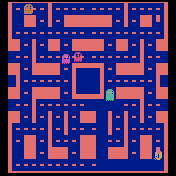}
    \parbox[t][\myheight]{\linewidth}{Move Down}
    \includegraphics[width=\linewidth]{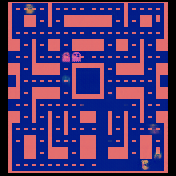}
    \includegraphics[width=\linewidth]{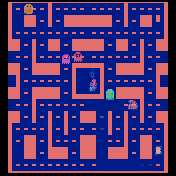}
    \end{minipage}
    %state 1
      \begin{minipage}{0.19\linewidth}
    \centering
    \parbox[t][\myheight]{\linewidth}{Move Right}
    \includegraphics[width=\linewidth]{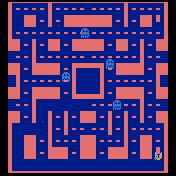}
    \parbox[t][\myheight]{\linewidth}{Move Down}
    \includegraphics[width=\linewidth]{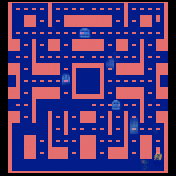}
    \includegraphics[width=\linewidth]{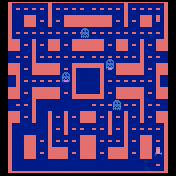}
    \end{minipage}
    % state 2
      \begin{minipage}{0.19\linewidth}
    \centering
    \parbox[t][\myheight]{\linewidth}{Move Up}
    \includegraphics[width=\linewidth]{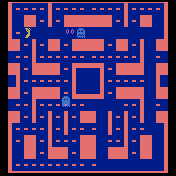}
    \parbox[t][\myheight]{\linewidth}{Move Down}
    \includegraphics[width=\linewidth]{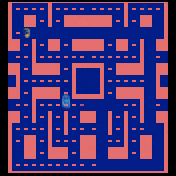}
    \includegraphics[width=\linewidth]{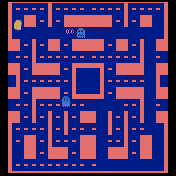}
    \end{minipage}
    % state 3
      \begin{minipage}{0.19\linewidth}
    \centering
    \parbox[t][\myheight]{\linewidth}{Do Nothing/Keep Direction}
    \includegraphics[width=\linewidth]{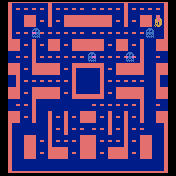}
    \parbox[t][\myheight]{\linewidth}{Move Down}
    \includegraphics[width=\linewidth]{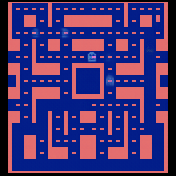}
    \includegraphics[width=\linewidth]{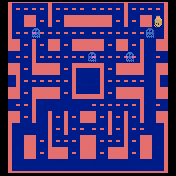}
    \end{minipage}
    % state 4
      \begin{minipage}{0.19\linewidth}
    \centering
    \parbox[t][\myheight]{\linewidth}{Move Up}
    \includegraphics[width=\linewidth]{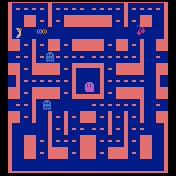}
    \parbox[t][\myheight]{\linewidth}{Move Right}
    \includegraphics[width=\linewidth]{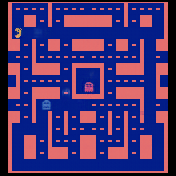}
    \includegraphics[width=\linewidth]{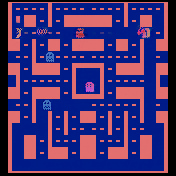}
    \end{minipage}
    \caption{Example counterfactual states for the \powerPillAgent{}.
    The first row shows the original states and the second and third rows show counterfactual states by \citet{olson2021} and our \starGAN{} approach respectively. 
    The states and actions are the same states that were used during our user study and were chosen by the HIGHLIGHTS-Div algorithm \cite{amir2018}.
    }
    \label{fig:power_pill_examples}
\end{figure*}

\begin{figure*}
    \small
    \centering
    \newcommand{\myheight}{0.9 \baselineskip} % textbox height
    % state 0
    \begin{minipage}{0.19\linewidth}
    \centering
    \parbox[t][\myheight]{\linewidth}{Move Right}
    \includegraphics[width=\linewidth]{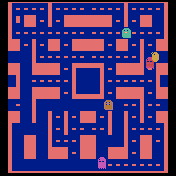}
    \parbox[t][\myheight]{\linewidth}{Move Left}
    \includegraphics[width=\linewidth]{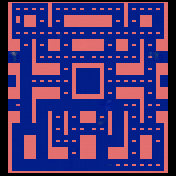}
    \includegraphics[width=\linewidth]{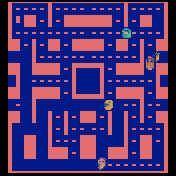}
    \end{minipage}
    %state 1
      \begin{minipage}{0.19\linewidth}
    \centering
    \parbox[t][\myheight]{\linewidth}{Move Down}
    \includegraphics[width=\linewidth]{figures/fear_ghost_agent/1_344_action4.png}
    \parbox[t][\myheight]{\linewidth}{Move Up}
    \includegraphics[width=\linewidth]{figures/fear_ghost_agent/1_344_action4_TargetDomain_1_Olson.png}
    \includegraphics[width=\linewidth]{figures/fear_ghost_agent/1_344_action4_TargetDomain_1_Stargan.png}
    \end{minipage}
    % state 2
      \begin{minipage}{0.19\linewidth}
    \centering
    \parbox[t][\myheight]{\linewidth}{Move Down}
    \includegraphics[width=\linewidth]{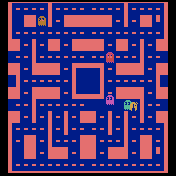}
    \parbox[t][\myheight]{\linewidth}{Move Up}
    \includegraphics[width=\linewidth]{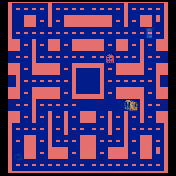}
    \includegraphics[width=\linewidth]{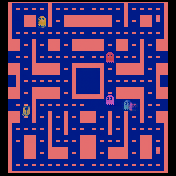}
    \end{minipage}
    % state 3
      \begin{minipage}{0.19\linewidth}
    \centering
    \parbox[t][\myheight]{\linewidth}{Move Right}
    \includegraphics[width=\linewidth]{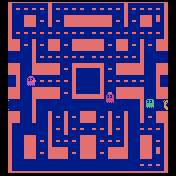}
    \parbox[t][\myheight]{\linewidth}{Move Left}
    \includegraphics[width=\linewidth]{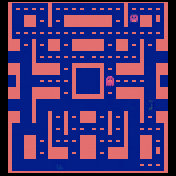}
    \includegraphics[width=\linewidth]{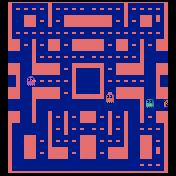}
    \end{minipage}
    % state 4
      \begin{minipage}{0.19\linewidth}
    \centering
    \parbox[t][\myheight]{\linewidth}{Move Right}
    \includegraphics[width=\linewidth]{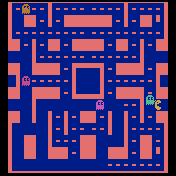}
    \parbox[t][\myheight]{\linewidth}{Move Left}
    \includegraphics[width=\linewidth]{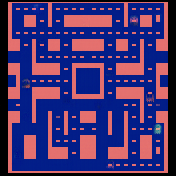}
    \includegraphics[width=\linewidth]{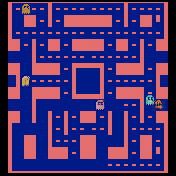}
    \end{minipage}
    \caption{Example counterfactual states for the \fearGhostAgent{}.
    The first row shows the original states and the second and third rows show counterfactual states by \citet{olson2021} and our \starGAN{} approach respectively. 
    The states and actions are the same states that were used during our user study and were chosen by the HIGHLIGHTS-Div algorithm \cite{amir2018}.
    }
    \label{fig:fear_ghost_examples}
\end{figure*}

\begin{figure*}
    \small
    \centering
    \newcommand{\myheight}{0.9 \baselineskip} % textbox height
    % state 0
    \begin{minipage}{0.19\linewidth}
    \centering
    \parbox[t][\myheight]{\linewidth}{Move Left}
    \includegraphics[width=\linewidth]{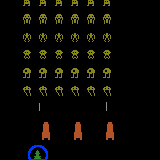}
    \parbox[t][\myheight]{\linewidth}{Right \& Fire}
    \includegraphics[width=\linewidth]{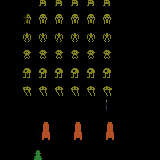}
    \includegraphics[width=\linewidth]{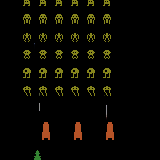}
    \end{minipage}
    %state 1
      \begin{minipage}{0.19\linewidth}
    \centering
    \parbox[t][\myheight]{\linewidth}{Right \& Fire}
    \includegraphics[width=\linewidth]{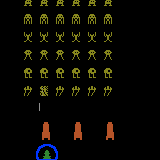}
    \parbox[t][\myheight]{\linewidth}{Move Left}
    \includegraphics[width=\linewidth]{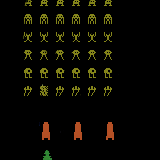}
    \includegraphics[width=\linewidth]{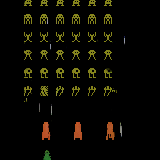}
    \end{minipage}
    % state 2
      \begin{minipage}{0.19\linewidth}
    \centering
    \parbox[t][\myheight]{\linewidth}{Left \& Fire}
    \includegraphics[width=\linewidth]{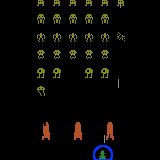}
    \parbox[t][\myheight]{\linewidth}{Move Right}
    \includegraphics[width=\linewidth]{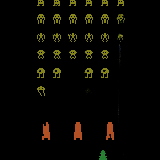}
    \includegraphics[width=\linewidth]{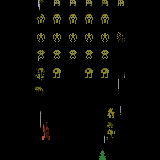}
    \end{minipage}
    % state 3
      \begin{minipage}{0.19\linewidth}
    \centering
    \parbox[t][\myheight]{\linewidth}{Right \& Fire}
    \includegraphics[width=\linewidth]{figures/ablated_spaceInvaders/original/3_13_action4.png}
    \parbox[t][\myheight]{\linewidth}{Move Left}
    \includegraphics[width=\linewidth]{figures/ablated_spaceInvaders/olson/CF_FromFile_3_13_action4_TargetDomain_3.png}
    \includegraphics[width=\linewidth]{figures/ablated_spaceInvaders/starGAN/CF_FromFile_3_13_action4_TargetDomain_3.png}
    \end{minipage}
    % state 4
      \begin{minipage}{0.19\linewidth}
    \centering
    \parbox[t][\myheight]{\linewidth}{Right \& Fire}
    \includegraphics[width=\linewidth]{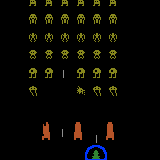}
    \parbox[t][\myheight]{\linewidth}{Move Left}
    \includegraphics[width=\linewidth]{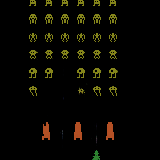}
    \includegraphics[width=\linewidth]{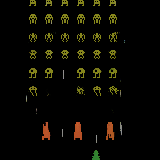}
    \end{minipage}
    \caption{Example counterfactual states for the flawed Space Invader agent.
    The first row shows the original states and the second and third rows show counterfactual states by \citet{olson2021} and our GANterfactual approach respectively. 
    The states were chosen by the HIGHLIGHTS-Div algorithm \cite{amir2018}.
    The counterfactual actions were chosen to be complete opposites of the original action. 
    Despite this big difference in the action, both approaches do not move the laser cannon (highlighted with a blue circle in the original frames) that the flawed agent does not see.
    }
    \label{fig:ablated_spaceInv}
\end{figure*}

\begin{figure*}
    \small
    \centering
    \newcommand{\myheight}{0.9 \baselineskip} % textbox height
    % state 0
    \begin{minipage}{0.19\linewidth}
    \centering
    \parbox[t][\myheight]{\linewidth}{Right \& Fire}
    \includegraphics[width=\linewidth]{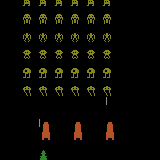}
    \parbox[t][\myheight]{\linewidth}{Move Left}
    \includegraphics[width=\linewidth]{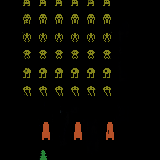}
    \includegraphics[width=\linewidth]{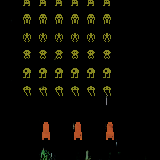}
    \end{minipage}
    %state 1
      \begin{minipage}{0.19\linewidth}
    \centering
    \parbox[t][\myheight]{\linewidth}{Right \& Fire}
    \includegraphics[width=\linewidth]{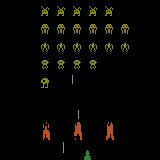}
    \parbox[t][\myheight]{\linewidth}{Move Left}
    \includegraphics[width=\linewidth]{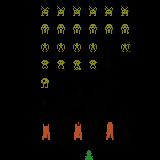}
    \includegraphics[width=\linewidth]{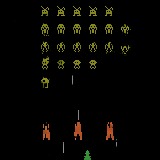}
    \end{minipage}
    % state 2
      \begin{minipage}{0.19\linewidth}
    \centering
    \parbox[t][\myheight]{\linewidth}{Right \& Fire}
    \includegraphics[width=\linewidth]{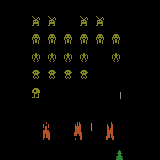}
    \parbox[t][\myheight]{\linewidth}{Move Left}
    \includegraphics[width=\linewidth]{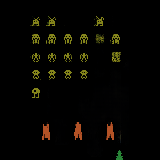}
    \includegraphics[width=\linewidth]{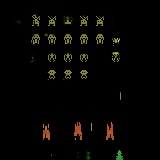}
    \end{minipage}
    % state 3
      \begin{minipage}{0.19\linewidth}
    \centering
    \parbox[t][\myheight]{\linewidth}{Left \& Fire}
    \includegraphics[width=\linewidth]{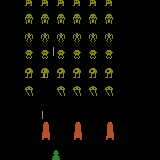}
    \parbox[t][\myheight]{\linewidth}{Move Right}
    \includegraphics[width=\linewidth]{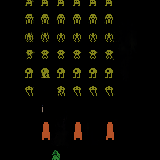}
    \includegraphics[width=\linewidth]{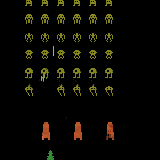}
    \end{minipage}
    % state 4
      \begin{minipage}{0.19\linewidth}
    \centering
    \parbox[t][\myheight]{\linewidth}{Right \& Fire}
    \includegraphics[width=\linewidth]{figures/spaceInvaders/original/4_7_action4.png}
    \parbox[t][\myheight]{\linewidth}{Move Left}
    \includegraphics[width=\linewidth]{figures/spaceInvaders/olson/CF_FromFile_4_7_action4_TargetDomain_3.png}
    \includegraphics[width=\linewidth]{figures/spaceInvaders/starGAN/CF_FromFile_4_7_action4_TargetDomain_3.png}
    \end{minipage}
    \caption{Example counterfactual states for the normal Space Invader agent.
    The first row shows the original states and the second and third rows show counterfactual states by \citet{olson2021} and our GANterfactual approach respectively. 
    The states are chosen by the HIGHLIGHTS-Div algorithm \cite{amir2018}.
    The counterfactual actions are chosen to be complete opposites of the original action. 
    In contrast to the counterfactuals for the flawed Space Invaders Agent, both approaches sometimes modify the laser cannon.
    }
    \label{fig:normal_spaceInv}
\end{figure*}

Figures \ref{fig:blue_ghost_examples}, \ref{fig:power_pill_examples}, \ref{fig:fear_ghost_examples}, \ref{fig:ablated_spaceInv}, and \ref{fig:normal_spaceInv} show example counterfactuals for both approaches tested in this work.

\section{Full User Study}
\begin{figure*}[tb] 
\centering
 \makebox[\textwidth]{\includegraphics[width=.9\paperwidth]{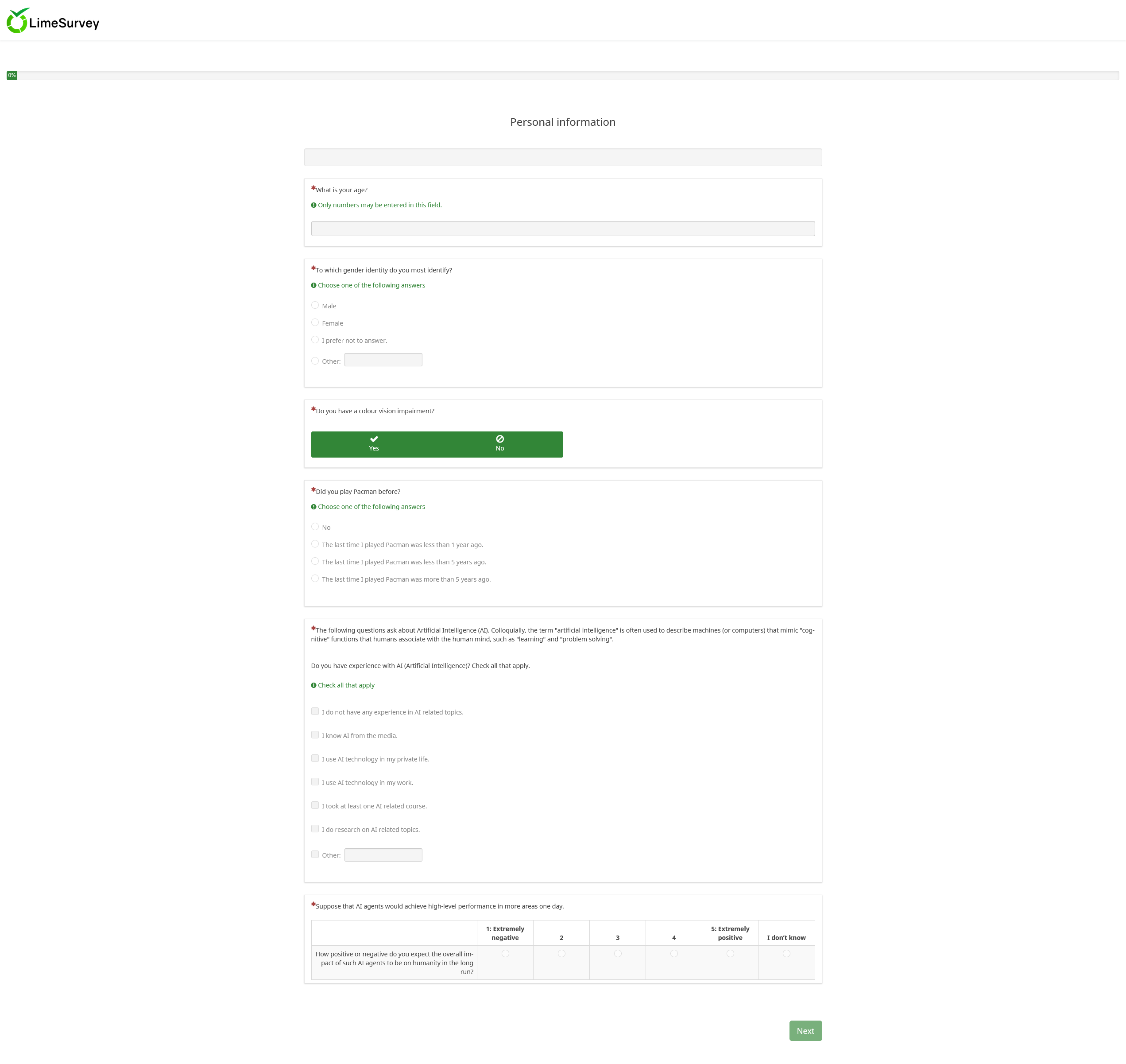}}
\caption{Demographic information.}
\label{fig:stdy_scr1}
\end{figure*}

\begin{figure*}[tb] 
\centering
 \makebox[\textwidth]{\includegraphics[width=.9\paperwidth]{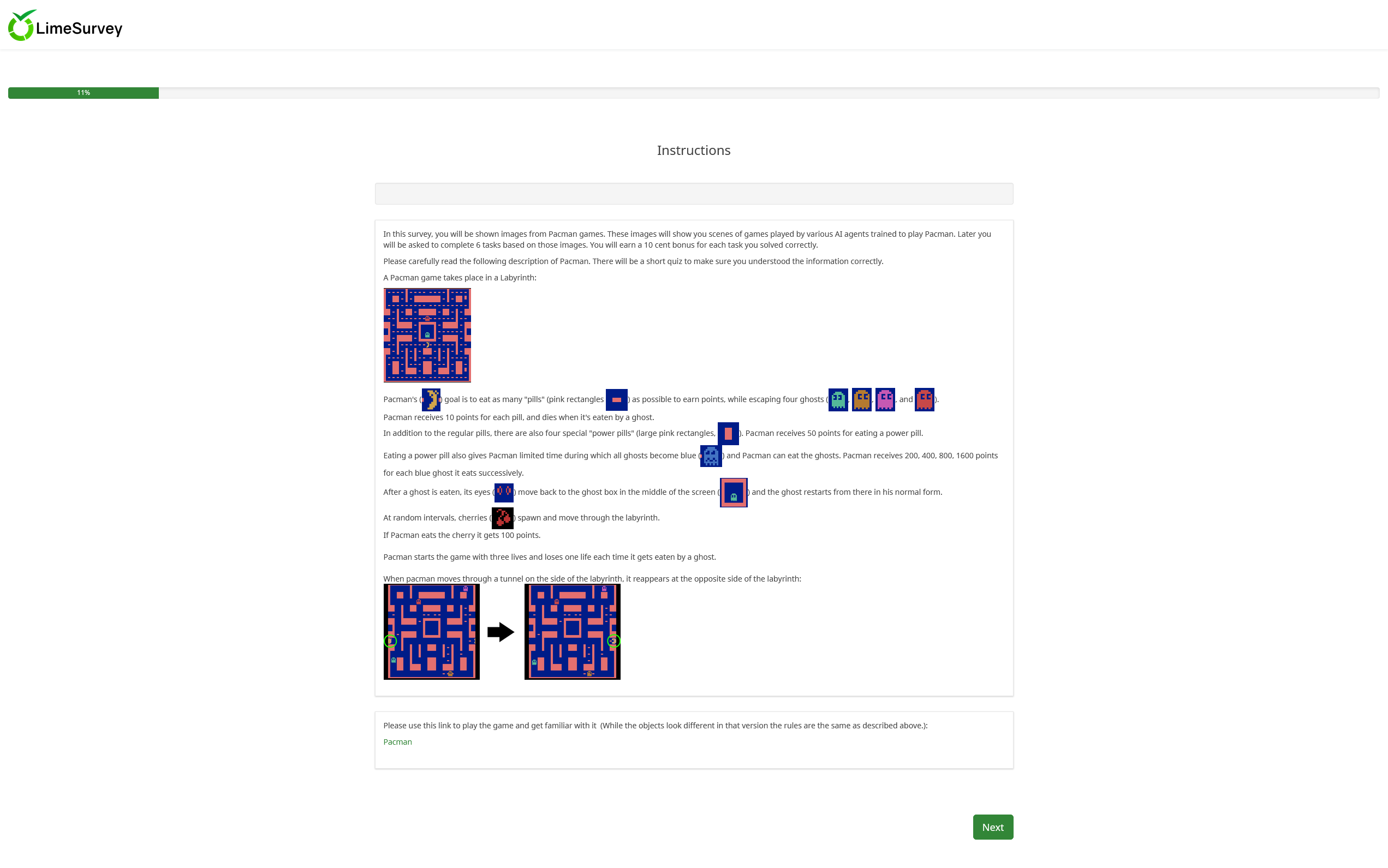}}
\caption{The Pacman tutorial.}
\label{fig:stdy_scr2}
\end{figure*}

\begin{figure*}[tb] 
\centering
 \makebox[\textwidth]{\includegraphics[width=.9\paperwidth]{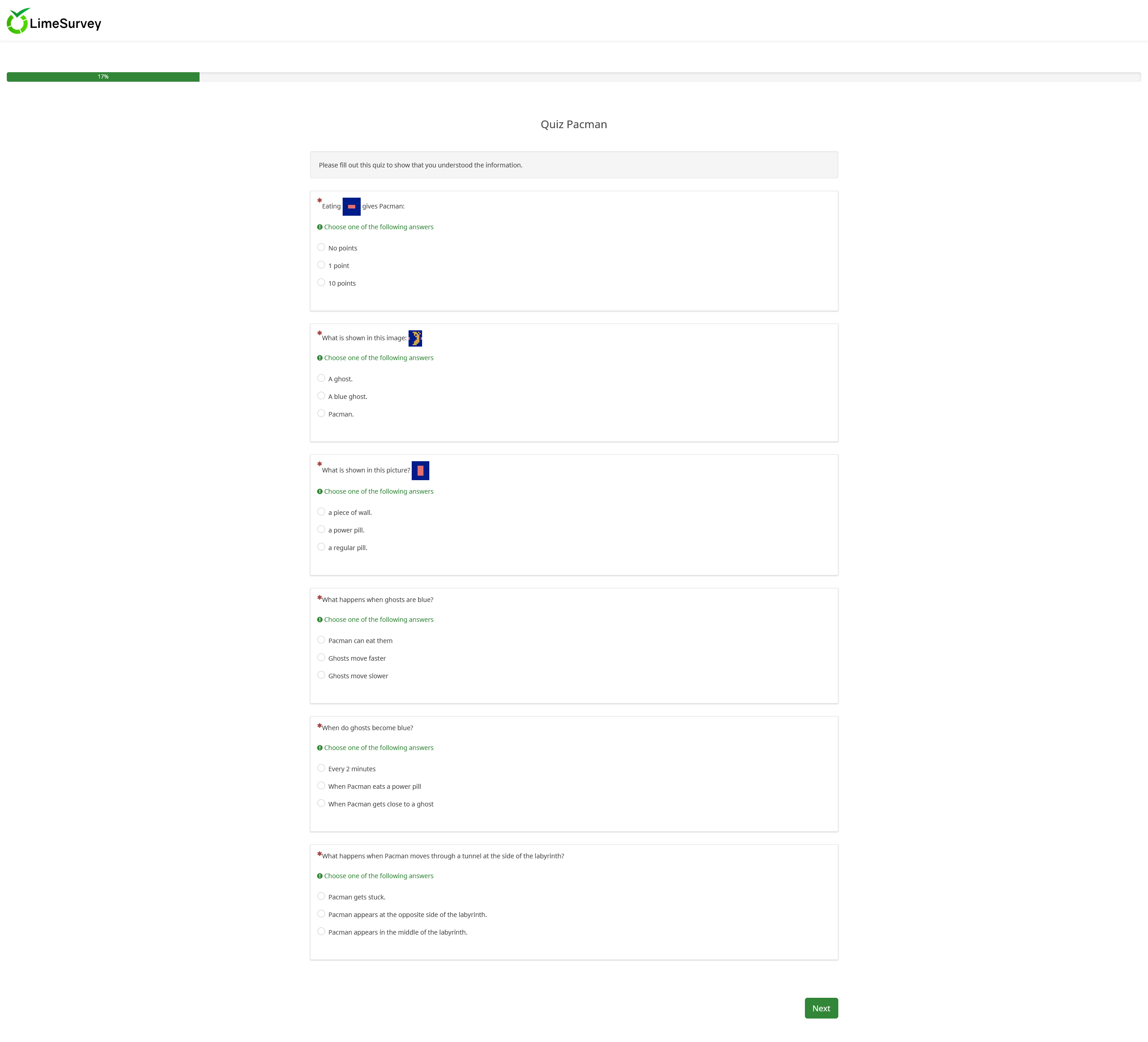}}
\caption{The Pacman quiz.}
\label{fig:stdy_scr3}
\end{figure*}

\begin{figure*}[tb] 
\centering
 \makebox[\textwidth]{\includegraphics[width=.9\paperwidth]{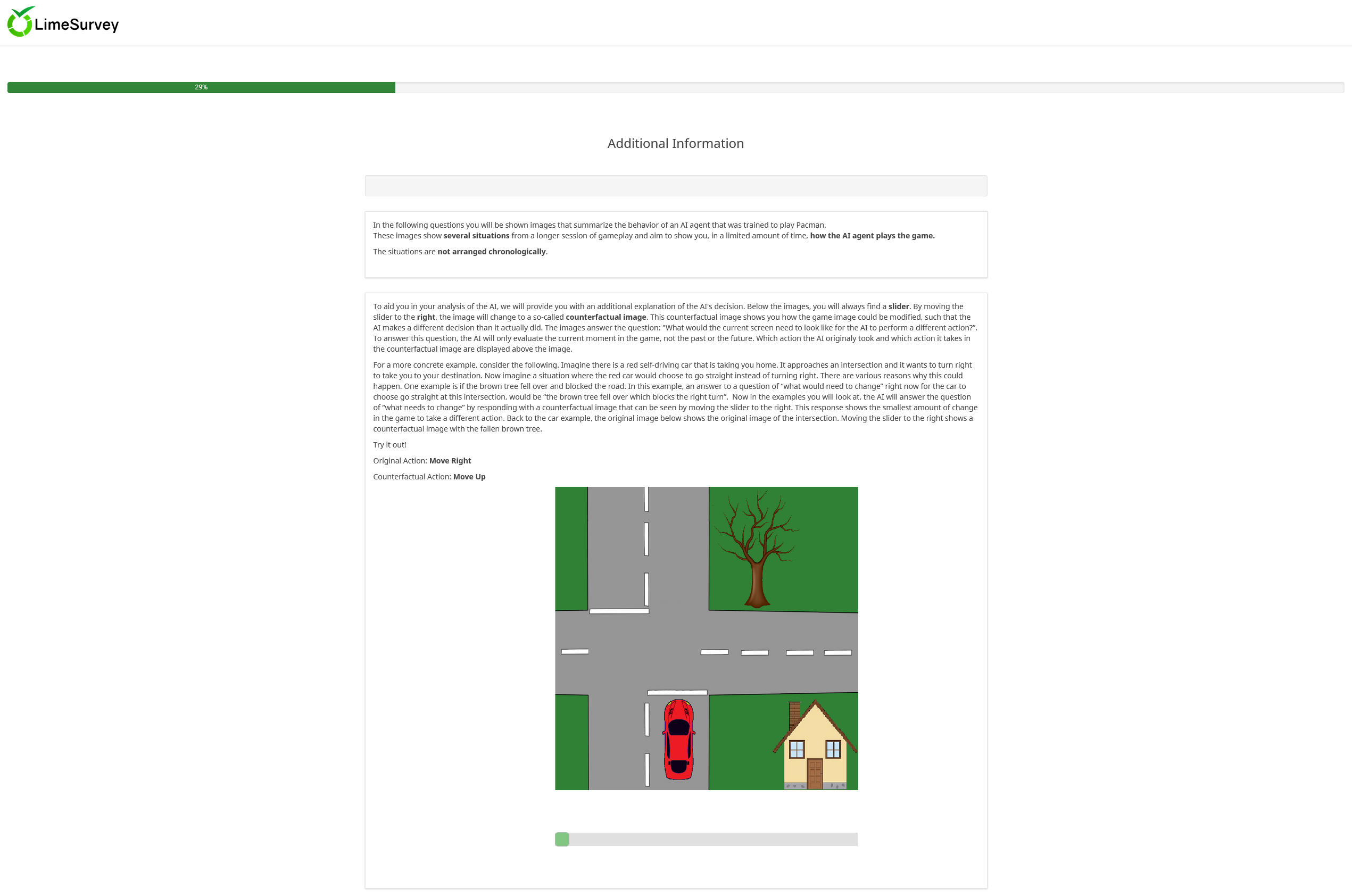}}
\caption{The first part of the counterfactual tutorial, which is built upon the tutorial by \citet{olson2021}.}
\label{fig:stdy_scr4_1}
\end{figure*}

\begin{figure*}[tb] 
\centering
 \makebox[\textwidth]{\includegraphics[width=.9\paperwidth]{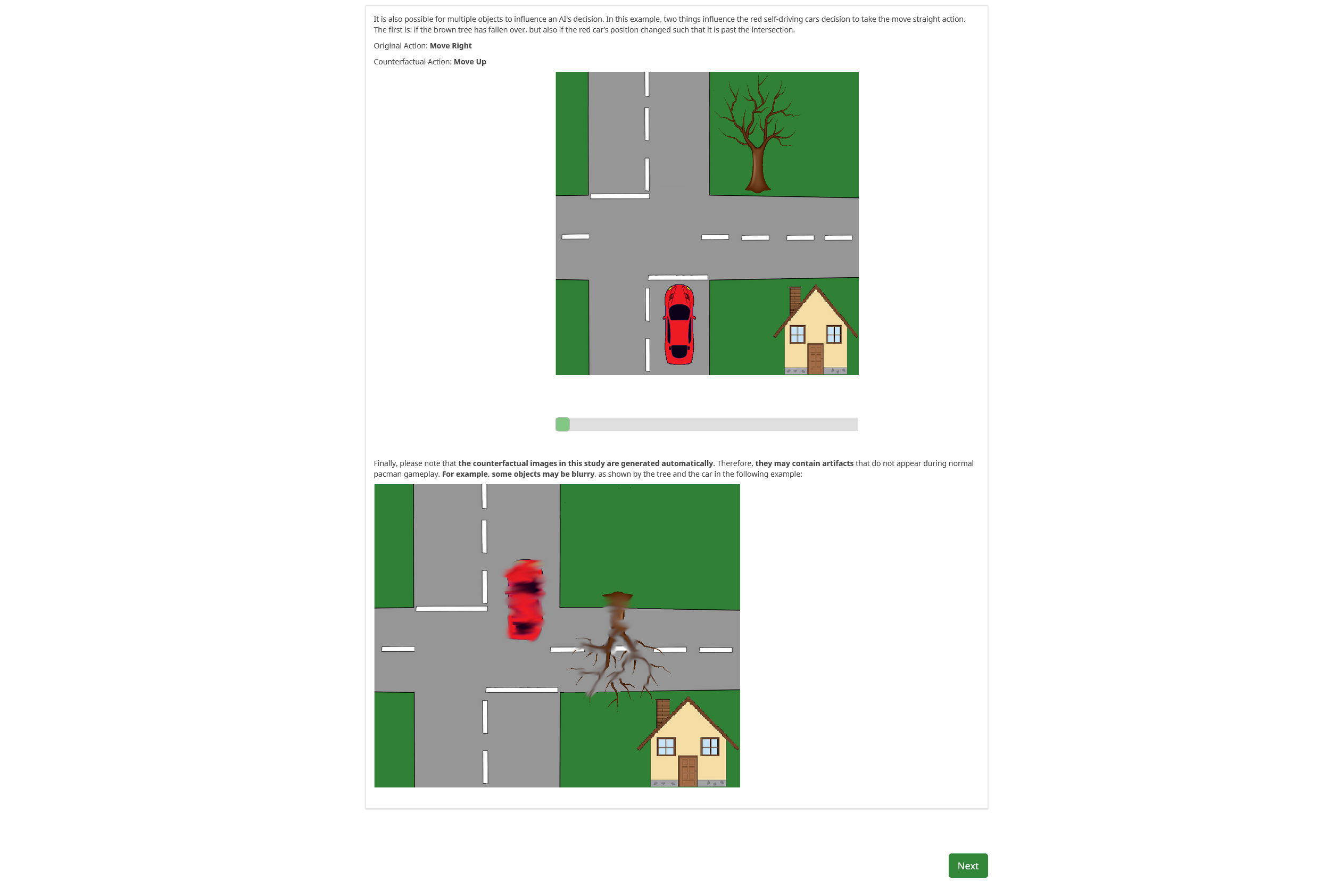}}
\caption{The second part of the counterfactual tutorial, which is built upon the tutorial by \citet{olson2021}.}
\label{fig:stdy_scr4_2}
\end{figure*}

\begin{figure*}[tb] 
\centering
 \makebox[\textwidth]{\includegraphics[width=.9\paperwidth]{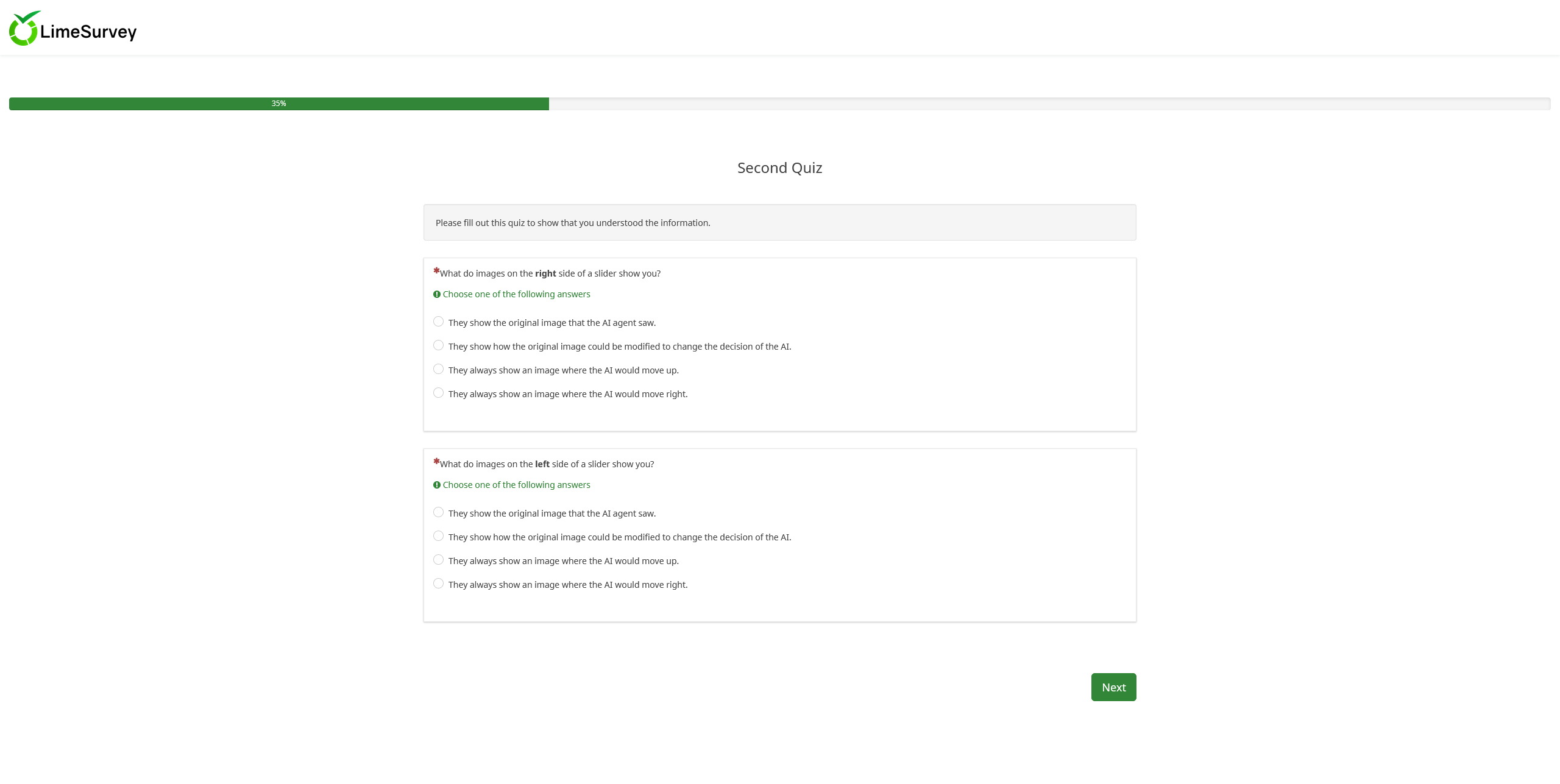}}
\caption{The quiz about counterfactual explanations.}
\label{fig:stdy_scr5}
\end{figure*}

\begin{figure*}[tb] 
\centering
 \makebox[\textwidth]{\includegraphics[width=.9\paperwidth]{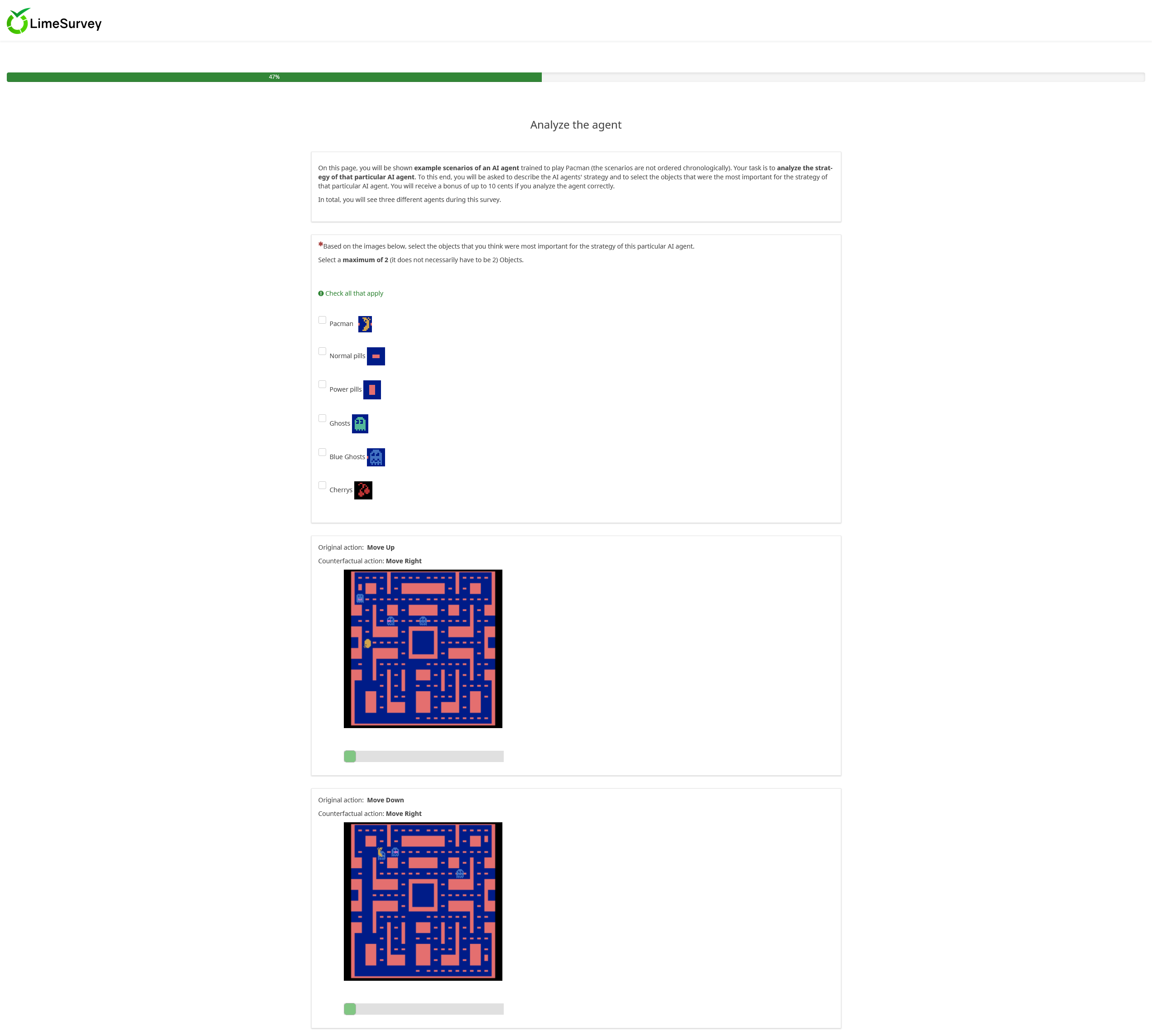}}
\caption{The first part of the \taskone{}. This task was repeated for all three agents. The order of the agents was randomized.}
\label{fig:stdy_scr6_1}
\end{figure*}

\begin{figure*}[tb] 
\centering
 \makebox[\textwidth]{\includegraphics[width=.9\paperwidth]{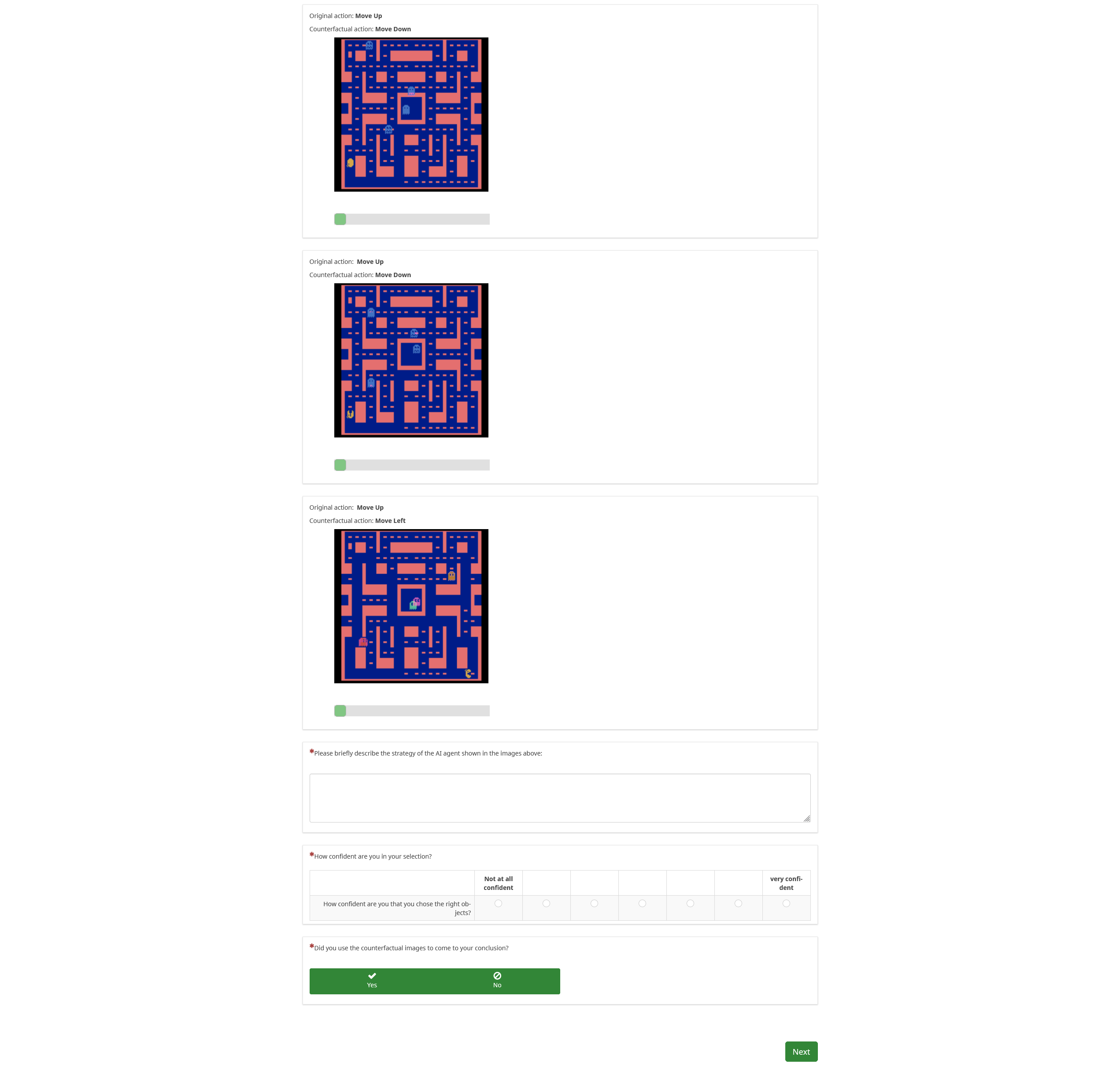}}
\caption{The second part of the \taskone{}. This task was repeated for all three agents. The order of the agents was randomized.}
\label{fig:stdy_scr6_2}
\end{figure*}

\begin{figure*}[tb] 
\centering
 \makebox[\textwidth]{\includegraphics[width=.9\paperwidth]{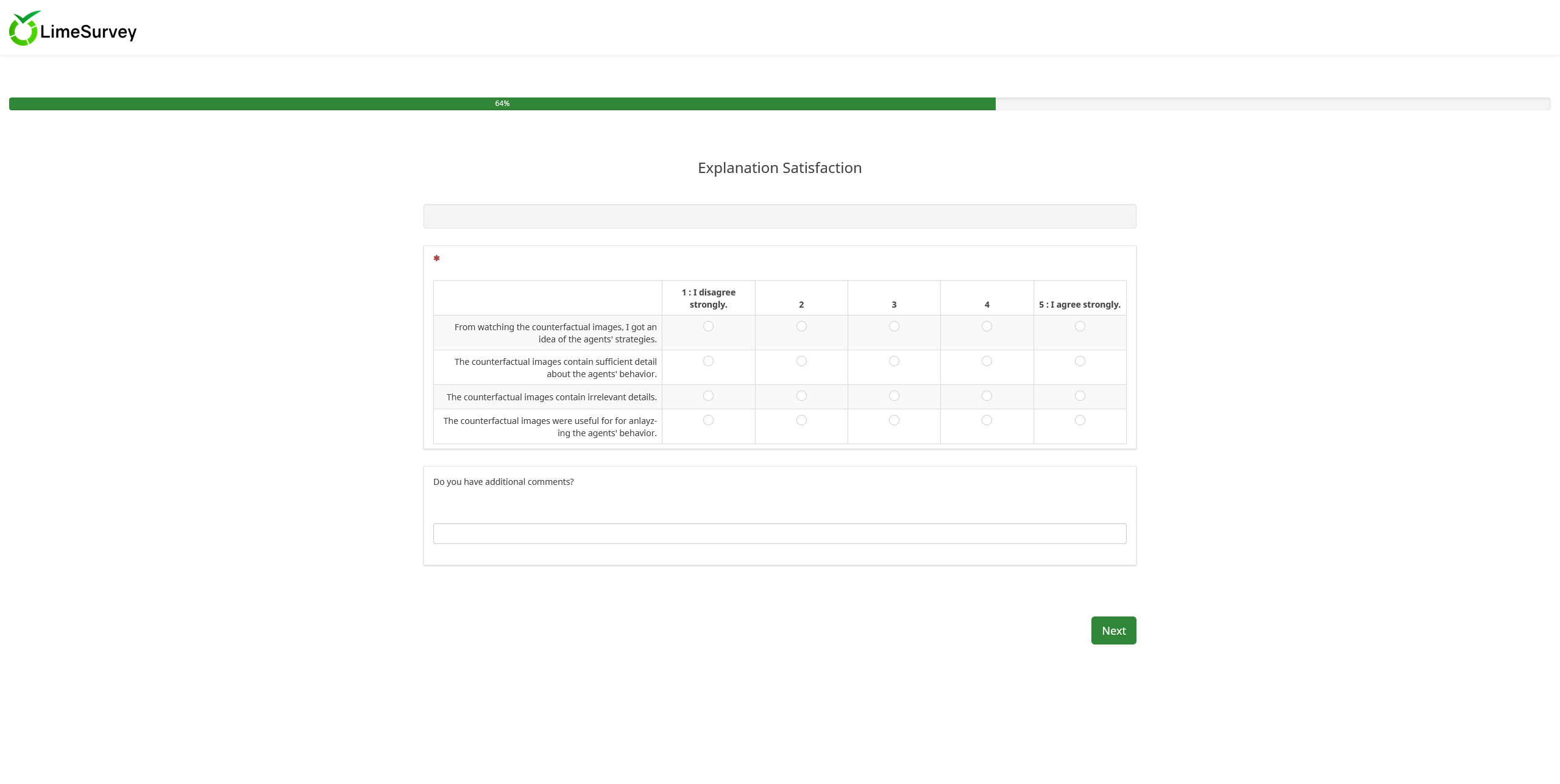}}
\caption{Explanation Satisfaction for the \taskone{}.}
\label{fig:stdy_scr9}
\end{figure*}

\begin{figure*}[tb] 
\centering
 \makebox[\textwidth]{\includegraphics[width=.9\paperwidth]{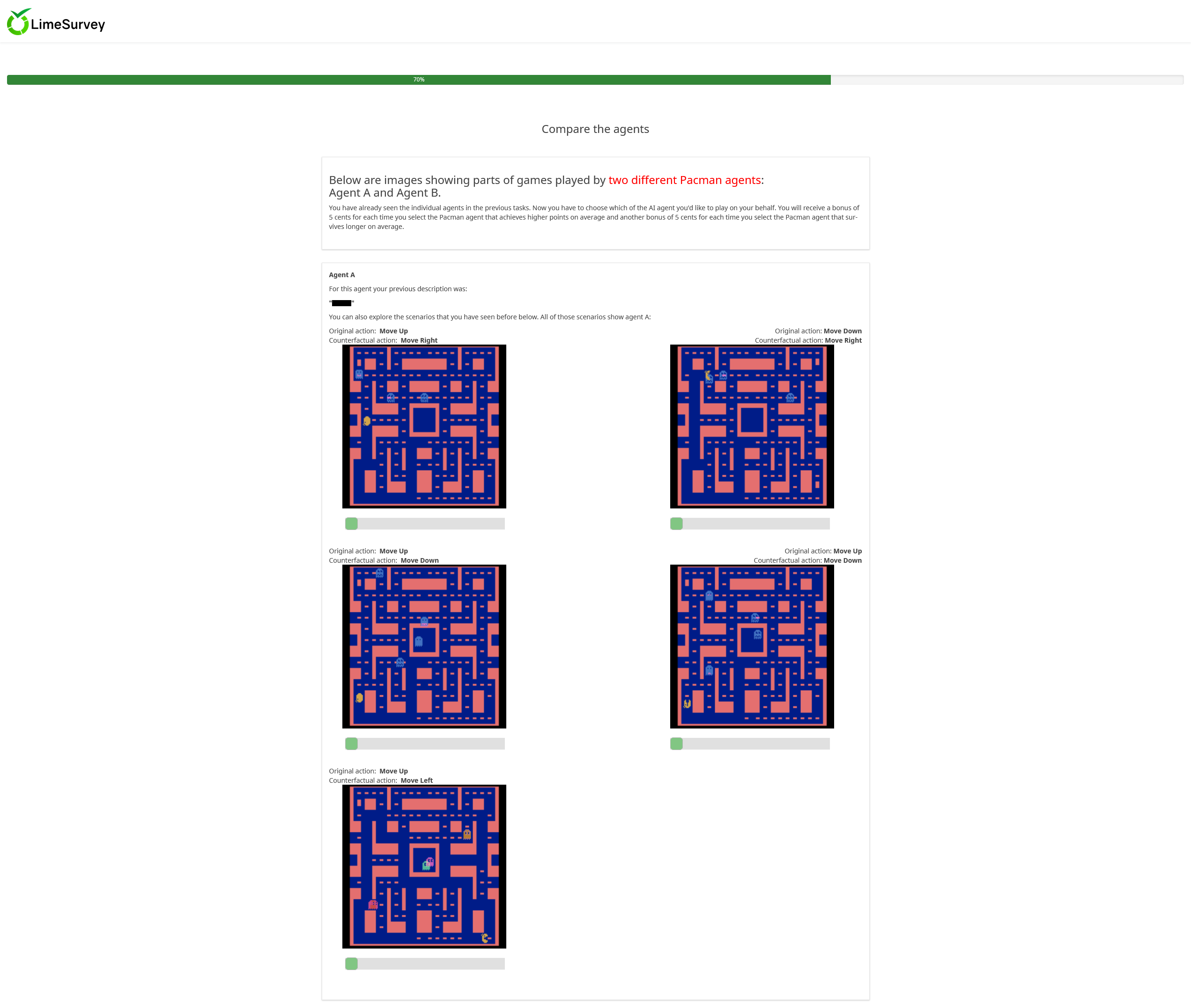}}
\caption{The first part of the \tasktwo{}. This task was repeated for all three agent pairs. The order of the pairs was randomized.}
\label{fig:stdy_scr10_1}
\end{figure*}

\begin{figure*}[tb] 
\centering
 \makebox[\textwidth]{\includegraphics[width=.9\paperwidth]{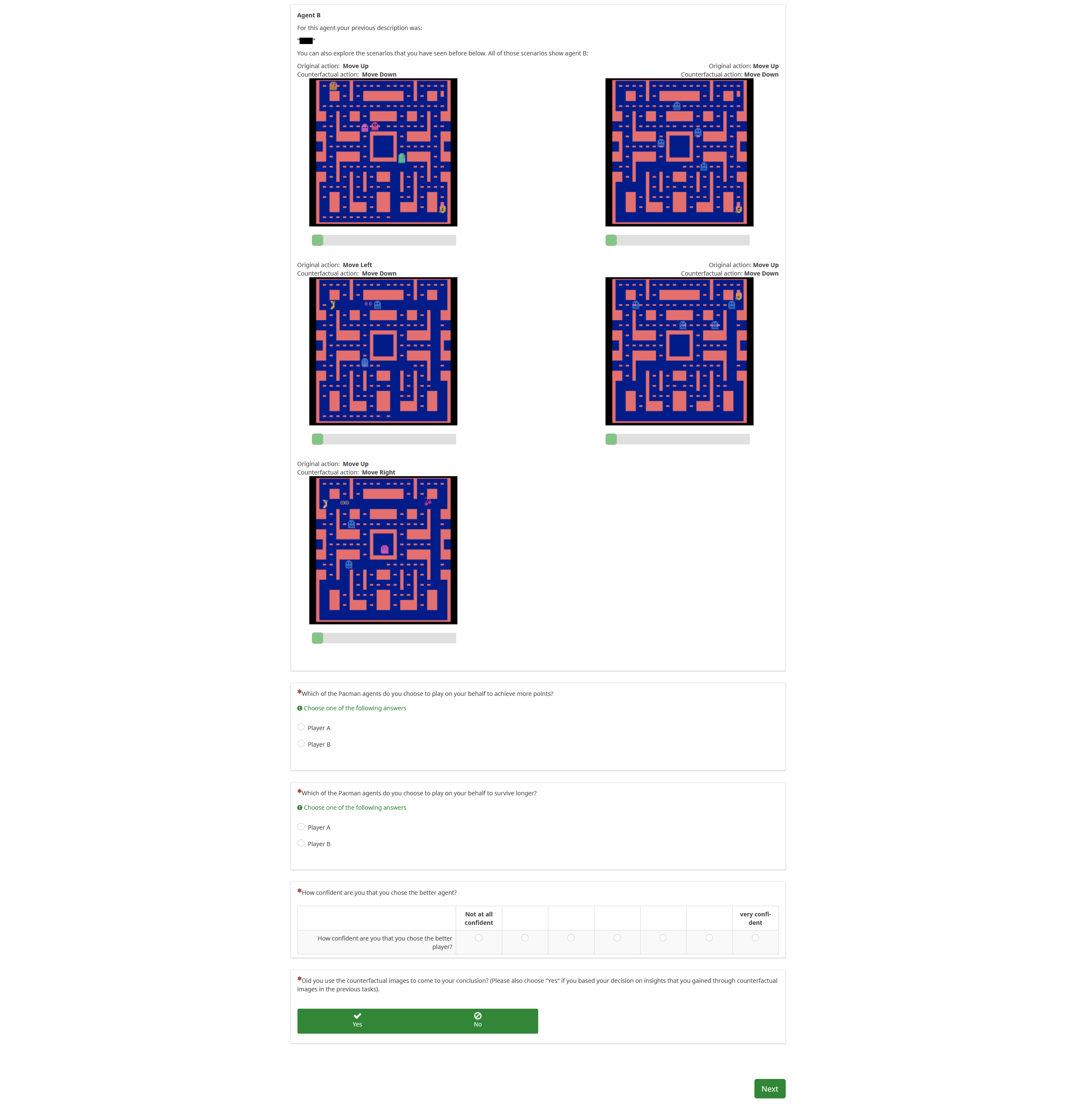}}
\caption{The second part of the \tasktwo{}. This task was repeated for all three agent pairs. The order of the pairs was randomized.}
\label{fig:stdy_scr10_2}
\end{figure*}

\begin{figure*}[tb] 
\centering
 \makebox[\textwidth]{\includegraphics[width=.9\paperwidth]{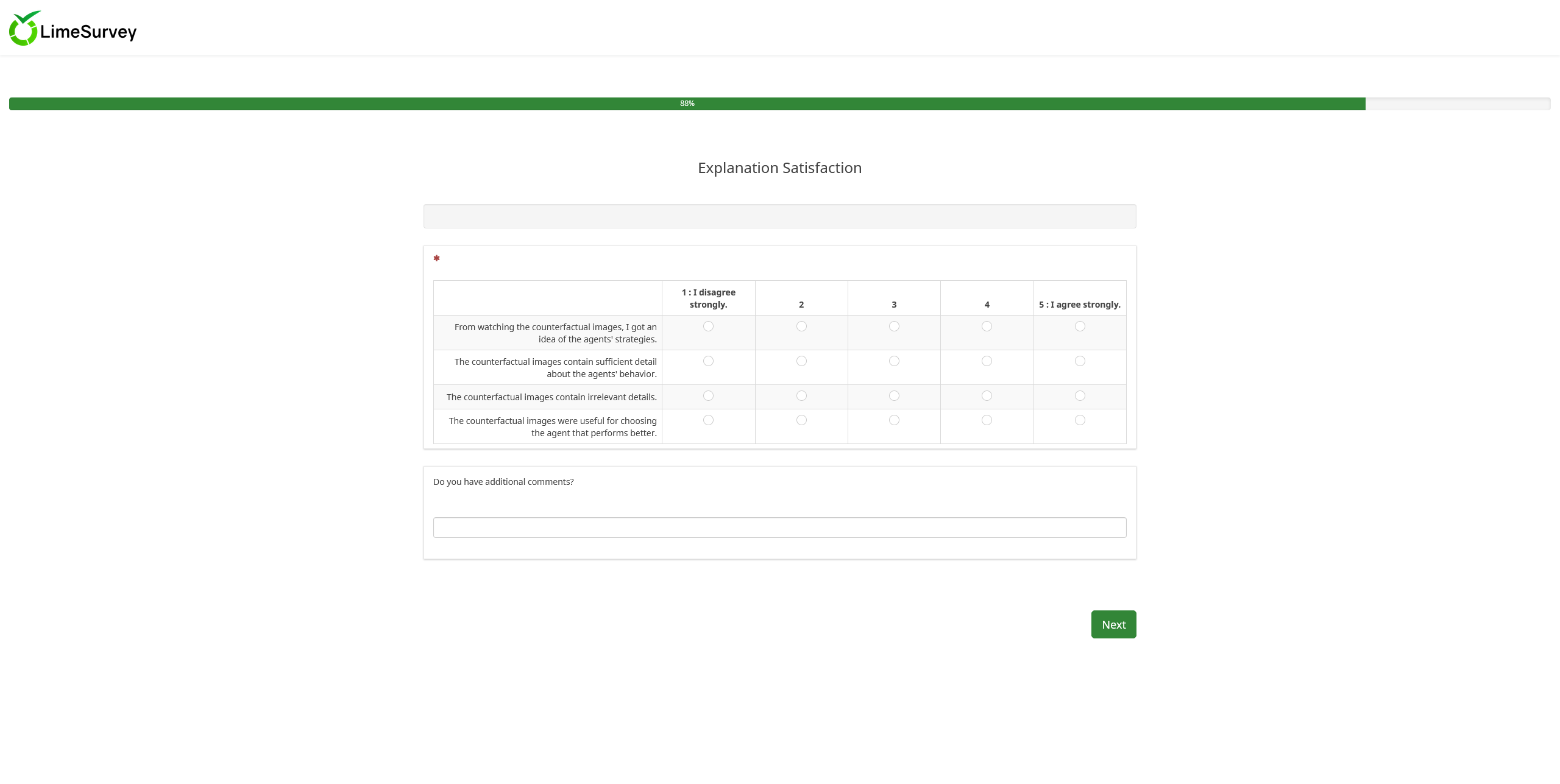}}
\caption{Explanation Satisfaction for the \tasktwo{}.}
\label{fig:stdy_scr13}
\end{figure*}

Figures \ref{fig:stdy_scr1} to \ref{fig:stdy_scr13} present screenshots of our user study. Exemplarily, we show the \olson{} condition.

% \appendix

% \input{sections/ap_results}

\end{document}